\pdfoutput=1
\documentclass[letterpaper, 10 pt, conference]{ieeeconf}
\IEEEoverridecommandlockouts                              
\let\labelindent\relax

\usepackage{mathptmx}

\usepackage{comment}

\usepackage{amsthm}
\usepackage{mathtools}      
\usepackage{amssymb}        
\usepackage{filecontents}
\usepackage{graphicx}       
\usepackage{marginnote}     
\usepackage{marvosym}       
\usepackage{overpic}        
\usepackage{tabularx}
\usepackage{cite}
\usepackage{color}
\usepackage[dvipsnames]{xcolor}
\usepackage[linesnumbered,algoruled,boxed,lined,noend]{algorithm2e}
\usepackage[normalem]{ulem}
\usepackage{epstopdf}
\usepackage{enumitem}
\usepackage[normalem]{ulem}
\usepackage{lipsum}

\usepackage{tikz}
\usepackage{tkz-euclide}
\usepackage{tikz-3dplot}
\usepackage{pgfgantt}

\usepackage[font=small]{caption}
\usepackage{subcaption}

\newif\ifdraft
\draftfalse

\ifdraft
\usepackage[paperheight=11in,paperwidth=9.5in,
			left=1.25in,right=1.25in,
			top=0.75in,bottom=0.75in,
			heightrounded,marginparwidth=1.2in,
			marginparsep=0.05in]{geometry}
\usepackage{xcolor}
\usepackage{xargs} 
\usepackage[textsize=footnotesize]{todonotes}
\newcommandx{\kg}[2][1=]{\todo[linecolor=red,
			backgroundcolor=red!10,bordercolor=red,#1]{KG: #2}}
\newcommandx{\jy}[2][1=]{\todo[linecolor=green,
			backgroundcolor=green!10,bordercolor=green,#1]{JY: #2}}
\newcommandx{\wt}[2][1=]{\todo[linecolor=yellow,
			backgroundcolor=yellow!10,bordercolor=yellow,#1]{WT: #2}}
\else
\newcommand{\kg}[1]{{}}
\newcommand{\jg}[1]{{}}
\newcommand{\jy}[1]{{}}
\newcommand{\wt}[1]{{}}
\fi

\newif\iftwocolumn
\twocolumntrue

\newtheorem{problem}{Problem}

\newtheorem{lemma}{Lemma}[section]

\newtheorem{corollary}{Corollary}[section]
\newtheorem{theorem}{Theorem}[section]
\theoremstyle{definition}
\newtheorem{definition}{Definition}[section]
\theoremstyle{remark}

\SetKwProg{Fn}{Function}{}{}
\SetKwComment{Comment}{$\triangleright$\ }{}

\makeatletter
\def\subsubsection{\@startsection{subsubsection}
                                 {3}
                                 {\z@ \hspace*{1mm}}
                                 {0ex plus 0.1ex minus 0.1ex}
                                 {0ex}
                                 {\normalfont\normalsize\itshape}}
\makeatother

\newcommand{\R}{\mathbb{R}}
\def\vrp{\textsc{VRP}\xspace}
\def\tsp{\textsc{TSP}\xspace}

\def\cvrp{\textsc{CVRP}\xspace}
\def\cvrg{\textsc{CVRG}\xspace}
\def\tpart{\textbf{\textsc{$3$-Partition}}\xspace}

\title{\Large \bf
Capacitated Vehicle Routing with Target Geometric Constraints 
}
\author{Kai~Gao\qquad\qquad
\qquad\qquad
Jingjin~Yu \vspace{-12mm}
\thanks{K. Gao and J. Yu are with the Department of Computer Science, 
Rutgers University at New Brunswick. 
E-mails: \{kai.gao, jingjin.yu\}@rutgers.edu.
This work is supported in part by NSF awards IIS-1734492, IIS-1845888, 
and CCF-1934924.}
}

\begin{document}

\maketitle


\begin{abstract}
We investigate the capacitated vehicle routing problem (\cvrp) under
a robotics context, where a vehicle with limited payload must complete 
delivery (or pickup) tasks to serve a set of geographically distributed 
customers with varying demands. In classical \cvrp, a customer location 
is modeled as a point. In many robotics applications, however, it is 
more appropriate to model such ``customer locations'' as 2D regions. 
For example, in aerial delivery, a drone may drop a package anywhere 
in a customer's lot. This yields the problem of \cvrg 
(\textbf{C}apacitated \textbf{V}ehicle \textbf{R}outing with Target 
\textbf{G}eometric Constraints). Computationally, \cvrp is 
already strongly NP-hard; \cvrg is therefore more challenging. 
Nevertheless, we develop fast algorithms for \cvrg, capable of 
computing high quality solutions for hundreds of regions. Our algorithmic 
solution is guaranteed to be optimal when customer regions are convex.  
Numerical evaluations show that our proposed methods significantly 
outperform greedy best-first approaches. 
Comprehensive simulation studies confirm the effectiveness of our 
methods.

\end{abstract}

\section{Introduction}\label{sec:intro}
The Capacitated Vehicle Routing Problem (\cvrp) naturally 
arises in many robotics applications, e.g., autonomous truck 
routing on road networks \cite{liu2003mixed}, aerial delivery 
\cite{wen2016multi,karak2019hybrid}, 
clutter removal \cite{TanYu19ISRR}, and so on. In \cvrp, a vehicle 
starting at a {\em depot} with limited load capacity is tasked to 
deliver (resp., pick up) goods to (resp., from) geographically scattered 
{\em customers}. The goal is to minimize the total distance that the 
vehicle must travel to complete all delivery/pickup tasks. \cvrp is 
closely related to the {\em vehicle routing problem} (\vrp) 
\cite{dantzig1959truck} and the {\em traveling salesperson problem} 
(\tsp) \cite{lawler1985traveling}, both of which are NP-hard problems 
that have been studied extensively \cite{toth2002vehicle,
garey1979computers}. In the literature, effective solutions to \cvrp 
have mainly been based on {\em branch-and-bound}, {\em branch-and-cut}, 
and related methods \cite{laporte1987exact,
miller1995matching,toth1997exact,caprara1997branch,
baldacci2004exact}. 

\begin{figure}[ht]
\vspace*{3mm}
\begin{center}
\begin{overpic}[width={\iftwocolumn \columnwidth \else 5in \fi},tics=5]
{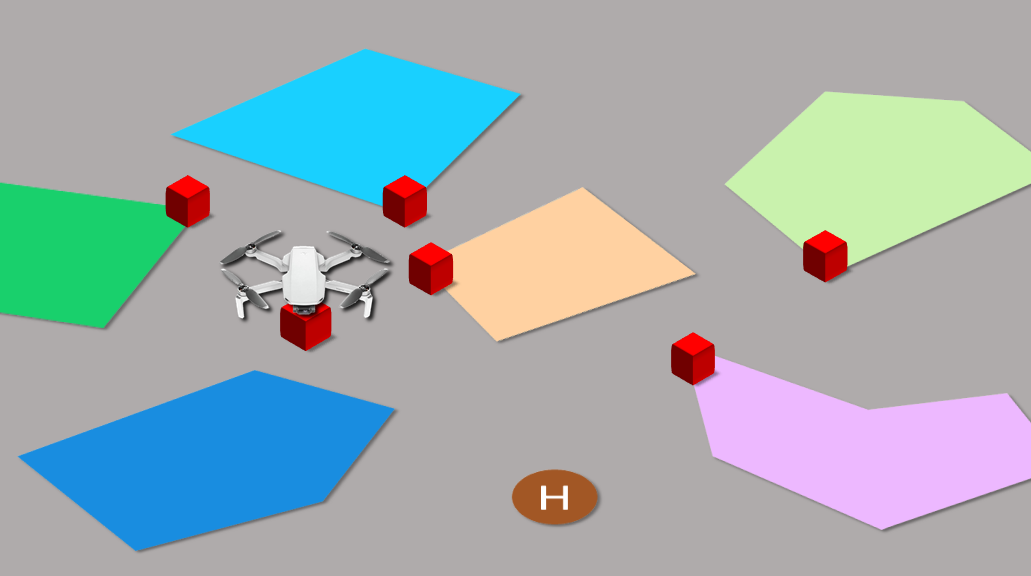}
\end{overpic}
\end{center}
\caption{\label{fig:drone}A \cvrg application where an aerial vehicle 
with limited load capacity is tasked to deliver 
packages to multiple regions. For a given region, the package may 
be dropped at any point inside the region. To make fast deliveries 
to all regions, the vehicle must carefully plan multiple tours with 
optimized delivery locations, which is computationally intractable. 
``H'' denotes the depot.}
\vspace{-3mm}
\end{figure}

In this work, we examine a variant of \cvrp when customer locations 
have non-trivial geometry and study the resulting optimality structure. 
In this \cvrp variant, instead of having a point location, each customer 
now occupies a contiguous 2D region. We denote the problem as 
\textbf{C}apacitated \textbf{V}ehicle \textbf{R}outing with (non-trivial) Target \textbf{G}eometric Constraints (\cvrg). \cvrg accurately 
models multiple real-world application scenarios. For example, in aerial 
delivery (or pickup), a parcel may be dropped (picked up) anywhere inside a given region 
(Fig.~\ref{fig:drone}). Similarly, optimal algorithms for \cvrg can provide
the most desirable solutions for helicopter-based rescuing missions where 
people are isolated on ``islands'' during floods or other natural disasters. 

Our algorithmic attack on the \cvrg problem seeks to decouple the 
\cvrp element and the additional constraint induced by non-trivial 
customer location geometry. 
On the \cvrp side, we examine existing state-of-the-art linear 
programming based solvers and develop a combinatorial algorithm that 
uses dynamic programming (DP) to chain together individual tours (a {\em
tour} is a round trip of the robot starting from the depot; generally,
multiple tours are required to solve a \cvrp).
%
On the geometric constraint side, we prove that optimizing the 
delivery/pickup locations over a tour with fixed customer sequence 
induces an optimization problem that can be solved efficiently 
when customer regions are convex, allowing us to construct 
efficient subroutines for computing the corresponding optimal tour. 
A hierarchical combination of the \cvrp subroutines and the 
geometric optimization subroutines prove to be highly effective
when compared with greedy best-first approaches, yielding solutions
with much higher quality. 
At the same time, we note that \cvrp is strongly NP-hard even when 
customer demand is lower bounded, which we prove via a reduction 
from \tpart \cite{garey1975complexity}. 
The effective algorithmic solutions developed for \cvrg, in particular 
an optimal solution method for convex customer regions and a fast,
high-quality finite horizon heuristic algorithm based on DP,  form 
the main contributions of this work. 
The algorithms developed in this paper are thoroughly evaluated in 
simulation studies, including a physics engine based drone delivery scenario, which confirm their effectiveness. 

Beside its relevance to \cvrp, \cvrg is closely related to research on 
object rearrangement in robotics. A diverse set of methods have been 
applied to tackle object rearrangement tasks, including search based 
approaches \cite{ota2009rearrangement} and symbolic reasoning based 
approaches \cite{havur2014geometric}. In contrast, in 
\cite{krontiris2015dealing,HanStiKroBekYu18IJRR, wang2021uniform, gaorunning}, more focus is put on 
taming the combinatorial explosion caused by the large number of objects. 
There, even the seemingly simple problem of rearranging unlabeled objects 
turns out to be NP-hard if an optimal solution is sought after 
\cite{HanStiKroBekYu18IJRR}, which echoes the computational challenge we 
face in the current study. 
From an application perspective, our work applies to scenarios 
including aerial delivery \cite{wen2016multi,karak2019hybrid}, 
disaster response \cite{pratt2013darpa,murphy2014disaster,TanYu19ISRR},
among others.
\cvrg can also model the truck behavior in the truck-drone collaborative 
delivery problems \cite{ha2018min}\cite{yoo2018drone}\cite{murray2015flying}. 
For each delivery, the truck only needs to reach the neighborhood of the 
customer and let the drone do the last-mile delivery.


\textbf{Organization}. The rest of the paper is organized as follows. 
In Section~\ref{sec:prelim}, \cvrg is stated, followed by a preliminary 
structural analysis and a hardness proof. In Section~\ref{sec:cvrpbdg}, 
we describe in detail our proposed algorithmic solutions for \cvrg,
which are thoroughly evaluated and compared in Section~\ref{sec:evaluation}.
We conclude in Section~\ref{sec:conclusion}.

\section{Preliminaries}\label{sec:prelim}
\vspace{-1mm}
\subsection{\cvrp with Target Geometric Constraints (\cvrg)}
\vspace{-1mm}
In a standard {\em capacitated vehicle routing problem} (\cvrp) 
\cite{christofides1969algorithm}, a robot (vehicle) with a fixed load 
capacity is tasked to transport goods to multiple customers from a 
depot. More formally, let $d \in \R^2$ be the location of a depot where 
a robot may carry a maximum load of $W$ and deliver different loads to 
a set of $n$ customers located at $c_1, \ldots, c_n$, respectively, with 
$c_i \in \R^2$ for $1 \le i \le n$. Each customer has a demand ``weight'' 
of $w_i$ that must be satisfied through delivery by the robot.\footnote{Because 
the absolute values of $W$ and $w_i$ do not matter, we set $W = 1$ 
in this paper, which means that $w_i \in (0, 1]$.}
The robot returns to the depot after all deliveries are finished. As 
a problem that is always feasible, the goal in solving \cvrp
is to minimize the total distance traveled by the robot. 
The problem of picking up goods is symmetrical to the problem of delivering 
goods. In presenting this work, we mainly use the delivery setup but note 
that the pickup setup is also used when appropriate and is equivalent. 

\cvrp generally requires a customer's demand to be met by a single 
delivery. This makes sense as each delivery operation itself will 
incur a non-trivial overhead (for both parties). In robotics 
operations, the overhead can be dropping of packages from air or 
grasping an object. We mention that the results that we develop
also apply if the environment is modeled as a graph, e.g., as a road 
network. 
When it comes to robotics applications, e.g., in aerial delivery, 
the customer location could have non-trivial geometry. We model such 
constraints by treating each delivery location as a simply-connected 
polygon. 

\vspace{-1mm}
\begin{problem}[\cvrp with Target Geometric Constraints 
(\cvrg)] Let $d \in \R^2$ be the location of a depot with unlimited 
supply. There are $n$ customers located in $P_1, \ldots, P_n$, where 
$P_i \subset \R^2$ is a simply-connected polygon. Customer 
$i$ has a demand of $w_i \in (0, 1]$, which is satisfied as a single 
supply of $w_i$ is delivered to any point $p \in P_i$. Find the 
minimum total distance required for a robot with unit capacity to 
complete all deliveries, starting from and ending at the depot. 
\end{problem}
\vspace{-1mm}

In \cvrg, it is not required that for $1 \le i, j \le n$, $P_i \cap 
P_j = \varnothing$, i.e., $P_i$ and $P_j$ may overlap. Whereas 
the non-overlapping case is suitable for applications like aerial 
delivery, the overlapping setup can be more suitable for applications 
like picking up objects which may fall on one another. We explicitly 
address the case where elements of $\{P_i\}$ overlap. 

\vspace{-1mm}
\subsection{Strong NP-Hardness of \cvrp/\cvrg}
\vspace{-1mm}
The classical \vrp problem is computationally intractable for large 
instances as the NP-hard \tsp readily reduces to it \cite{toth2002vehicle}.  
The same can be shown for \cvrp using similar arguments. 
However, solvers exist that can solve very large instances of \tsp 
near optimally very fast. \cvrp, on the other hand, proves much more 
challenging. We observe that \cvrp is in fact strongly NP-hard, even
when customer demand is lower bounded, i.e., the customer's demand is 
no smaller than $1/k$ for some integer $k$. Setting $k$ to be $4$ 
is sufficient to show strong NP-hardness.To show this, we first introduce the 
strongly NP-hard \tpart problem \cite{garey1975complexity}. 

\noindent
PROBLEM: \tpart\\
INSTANCE: A finite set A of $3m$ elements, a bound $B\in \mathbb{Z^+}$, 
and a ``size'' $s(a)\in \mathbb{Z^+}$ for each $a\in A$,
such that each $s(a)$ satisfies $B/4 < s(a) <B/2$ and $\sum_{a\in A} 
s(a) = mB$.\\
QUESTION: Is there a partition of $S$ into $m$ disjoint subsets $S_1, 
\ldots, S_m$ such that for $1\leq i\leq m$, 
$\sum_{a\in S_i} s(a) = B$?

\begin{theorem}
\cvrp is strongly NP-hard. 
\end{theorem}
\vspace{-1mm}
\begin{proof}Given a \tpart instance with $3m$ elements, we construct a
\cvrp instance as follows. In $\R^2$, let $d = (0, 0)$ be the depot and 
let there be $3m$ customers all located close to $(1, 0)$. That is, for 
any customer $i$, $1 \le i \le 3m$, $c_i \in B_{\varepsilon}(1, 0)$, which
is the $\varepsilon$ ball around the point $(1, 0)$ for some positive 
$\varepsilon \ll 1/m$. Let the robot have a unit capacity and let customer 
$i$ have a demand of $s(a_i)/B$. Since $s(a) > B/4$, each delivery can 
supply at most $3$ customers. 
To show that \cvrp is NP-hard, we will show that the \tpart problem admits 
an optimal partition if and only if the \cvrp problem admits a total travel 
distance of no more than $2m + 6m\varepsilon$. 

For the ``only if'' part, if the \tpart problem admits a partition of the $3m$ 
elements into $m$ sets $S_1, \ldots, S_m$ of three elements each such that 
$\sum_{a\in S_i} s(a) = B$, then a robot can complete all deliveries using 
a total of $m$ tours based on the partition. For the $i$-th tour, 
the robot may travel in a straight line to a customer in $S_i$, which incurs 
a distance of no more than $1 + \varepsilon$. Then, the robot will travel 
along straight lines to the other two customers in $S_i$, one after the 
other. These tours will incur a cost of no more than $2\varepsilon$ each. 
Finally, the robot can return to the depot with a distance of no more than 
$1 + \varepsilon$. The total cost in then no more than $2m + 6\varepsilon$. 

For the ``if'' part, if the \cvrp problem admits a solution with a cost 
of no more than $2m + 6m\varepsilon$ Since $\varepsilon \ll 1/m$, $2m + 
6m\varepsilon < 2m + 1$ and the robot can only make no more than $m$ tours 
from the depot and return. Since the robot can move at most a unit 
of supply to the customers per round tour, the robot can move at most a total 
supply of $m$. As the \cvrp problem is solved, this means that the robot 
must complete delivery to exactly $3$ customers since (1) partial deliveries
are not allowed and (2) at most three customers' demand can be fulfilled 
with a unit of supply. 
\end{proof}
\vspace{-1mm}

With \cvrp (and subsequently, \cvrg) being strongly NP-hard, it 
does not admit an FPTAS (fully polynomial-time approximation scheme)
unless P $=$ NP \cite{vazirani2013approximation}. That is, it is unlikely 
that efficient algorithms exist for solving \cvrp/\cvrg approximately 
optimally. 
\vspace{-1mm}


\section{Optimally Solving \cvrg}\label{sec:cvrpbdg}
\def\eia{\textsc{Elastic-Improv}\xspace}
\def\lia{\textsc{Local-Improv}\xspace}

\def\cvrgdp{\textsc{DP-CVRG}\xspace}
\def\cvrgdps{\textsc{DP-CVRG}\xspace}
\def\cvrgdpfh{\textsc{DP-FH-CVRG}\xspace}
\def\cvrgdpfhs{\textsc{FH}\xspace}
\def\cvrggd{\textsc{GD-CVRG}\xspace}
\def\cvrggds{\textsc{GD}\xspace}
\def\cvrgbcps{\textsc{BCP-CVRG}\xspace}

\def\cvrpdp{\textsc{DP-CVRP}\xspace}
\def\cvrpdps{\textsc{DP-CVRP}\xspace}
\def\cvrpdpfh{\textsc{FH-DP-CVRP}\xspace}
\def\cvrpdpfhs{\textsc{FH-CVRP}\xspace}
\vspace{-1mm}
Our algorithmic solution for \cvrg has two components: a \emph{dynamic 
programming} (DP) based algorithm for \cvrp (Section.~\ref{subsec:cvrpdp}) 
and a geometric optimization subroutine (Section.~\ref{subsec:subtour}) 
that efficiently compute optimal single tours in the presence of the geometric 
constraint. 
Our DP algorithm for \cvrg is guaranteed to be optimal when the customer
regions are convex. 
We also examine the relevant scenario when regions represent physical objects 
to be picked up and the objects have overlaps (Section.~\ref{subsec:overlap}). 

\vspace{-2mm}
\subsection{Exact and Finite-Horizon DP Algorithms for \cvrp}\label{subsec:cvrpdp}
\vspace{-1mm}
\noindent\textbf{Exact combinatorial algorithm}.
Our exact algorithm for \cvrp proceeds in two phases. In the first phase, 
we exhaustively compute the optimal cost $c_s$ of each valid tour starting 
from the depot that visits a subset of customers $s$. It is easy to see 
that there are at most $2^n$ tour combinations, many of which will be 
invalid as they will exceed the capacity limit. For each valid tour 
combination, the optimal cost is computed using an exact \tsp solver\cite{bellman1962dynamic}. The costs are then stored for the 
second phase of the computation. 

In the second phase, to select the optimal set of disjoint tours, 
a straightforward application of dynamic programming (DP) is used. 
Let $C$ be the set of all customers. Let $I \subset C$ be the set of 
customers that have not been served and let $S_I$ be the set of customer subsets of $I$ that can be visited in a single tour without violating the capacity constraint. Let $J_I$ be the optimal cost (i.e., the 
minimum total path length) to satisfy the demands of all of $I$, the 
standard DP recursion is 

\vspace{-2mm}
\begin{align}
\label{eq:cvrpdp}
J_I = \min_{s \in S_I}(J_{I-s} + c_s). 
\end{align}

\vspace{-2mm}
Note that $c_s$ is provided by the first phase through a direct look-up.
The DP algorithm provides significant computational savings by storing 
the optimal cost of all possible $I$'s. For $n$ customers, there are 
at most $2^n$ such $I$'s. The algorithm, denoted \cvrpdp, is straightforward 
to implement: we simply create a large enough table to hold the $2^n$
entries and then iteratively populate the entries. 

\noindent\textbf{Algorithm analysis}. In the first phase, there are at 
most $2^n$ tours to examine. Since each tour visits no more than $n$ 
elements, obtaining the optimal cost of a tour takes $O(T(n))$ 
time, where $T(n)=O(n^22^n)$ with a dynamic programming TSP solver. 
The first phase then takes time $O(n^2 4^n)$.

In the second phase, the incremental DP computation needs 
to go through $2^n$ possible $I$'s, starting from $I = \varnothing$ 
and eventually reach $I = C$. For each $I$, $S_I$ contains $O(2^{|I|})$ 
potential tours that need to be checked. The second phase then 
requires $O(2^n 2^n)$. Therefore, the overall computational complexity of \cvrpdp is then bounded 
by $ O(n^2 4^n)$. 

With the application of DP cutting down the computation from a naive 
$O(n!f(n))$ (where $f(n)$ is some polynomial function of $n$ for 
computing the optimal cost of a single sequence) to $O(n^2 4^n)$, 
significantly larger \cvrp instances can be solved optimally. As we 
will demonstrate, \cvrpdp is exact. More importantly, \cvrpdp 
readily allows the integration of additional geometric constraints 
in the \cvrg formulation.

\textbf{Finite-horizon heuristic}. While \cvrpdp computes exact 
solutions with decent performance, the computation time is still 
exponential with respect to $n$. To address this issue, we introduce 
a finite horizon heuristic which restricts the number of customers 
examined at any given time. That is, some fixed $h$ ($h \le n$) customers 
are selected on which \cvrpdp is run. Because high quality solutions to 
\cvrp are generally clustered (see, e.g., Fig.~\ref{fig:setup} and 
Fig.~\ref{fig:cvrgExample}), these $h$ 
candidates are selected in a way to facilitate such clustering. 
%
From running \cvrpdp on the $h$ candidates, we pick two most "convenient" tours with 
the least average cost (path length divided by the number of served customers), serve these customers, and repeat the process on 
the remaining customers. The heuristic algorithm, which we denote as 
\cvrpdpfh, has an apparent complexity of $O(n h^2 4^h)$, which is polynomial. 

\vspace{-1mm}
\subsection{Optimal Subtours Crossing Multiple Regions}\label{subsec:subtour}
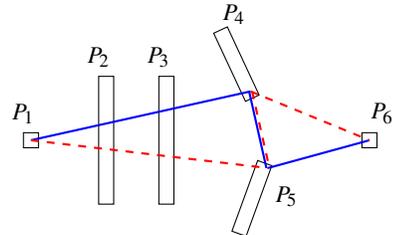
\begin{figure}[h!]
\vspace{-1mm}
\begin{center}
\begin{tikzpicture}
	 \coordinate (v1) at (-0.1, -0.1);
	 \coordinate (v2) at (0.9, 0.1);
	 \coordinate (v3) at (1.6, 0.1);
	 \coordinate (v4) at (0.1, 0.1);

	 \draw[black] let \p1=(v1), \p4=(v4) in [cm={cos(-90) ,-sin(-90) ,sin(-90) ,cos(-90) ,(-1 cm, 0.75 cm)}] 
	      (\p1) rectangle (\p4) node[anchor=south] {\small{$P_1$}};
	 \draw[black] let \p1=(v1), \p3=(v3) in [cm={cos(-90) ,-sin(-90) ,sin(-90) ,cos(-90) ,(0 cm,0 cm)}] 
	      (\p1) rectangle (\p3) node[anchor=south] {\small{$P_2$}};
	 \draw[black] let \p1=(v1), \p3=(v3) in [cm={cos(-90) ,-sin(-90) ,sin(-90) ,cos(-90) ,(0.8 cm,0 cm)}] 
	      (\p1) rectangle (\p3) node[anchor=south] {\small{$P_3$}};
	 \draw[black] let \p1=(v1), \p2=(v2) in [cm={cos(-70) ,-sin(-70) ,sin(-70) ,cos(-70) ,(1.8 cm,-0.4 cm)}] 
	      (\p1) rectangle (\p2);
  \node at (2.4, 0) {\small{$P_5$}};
	 \draw[black] let \p1=(v1), \p2=(v2) in [cm={cos(-115) ,-sin(-115) ,sin(-115) ,cos(-115) ,(1.9 cm,1.4 cm)}] 
	      (\p1) rectangle (\p2) node[anchor=south west] {\small{$P_4$}};
	 \draw[black] let \p1=(v1), \p4=(v4) in [cm={cos(-90) ,-sin(-90) ,sin(-90) ,cos(-90) ,(3.5 cm, 0.75 cm)}] 
	      (\p1) rectangle (\p4) node[anchor=south west] {\small{$P_6$}};

   \draw[thick,color=blue,-] (-1,0.75) -- (1.92, 1.4); 
	 \draw[thick,dashed,color=red,-] (-1, 0.75) -- (2.15, 0.38);
   \draw[thick,color=blue,-] (1.9, 1.4) -- (2.13, 0.38);
	 \draw[thick,dashed,color=red,-] (1.94, 1.4) -- (2.17, 0.38);
   \draw[thick,color=blue,-] (2.15, 0.38) -- (3.5, 0.75); 
	 \draw[thick,dashed,color=red,-] (1.92, 1.4) -- (3.5, 0.75);

  \end{tikzpicture}
\end{center}
\vspace{-2mm}
\caption{\label{fig:no-dp} Dynamic programming property existing in TSP tour 
computation fails to hold as the vertices become polygons.}
\vspace{-2mm}
\end{figure}

\vspace{-1mm}
For the first phase of \cvrpdp, the optimal cost for a single tour
 is computed by solving a TSP with DP, where a certain DP
property holds: given five customers $c_1$ to $c_5$, and two 
paths $c_1c_2c_3c_4c_5$ and $c_1c_2c_3c_5c_4$, the optimal cost of 
$c_1c_2c_3$ is the same in both and the computation for this part 
only needs to be done once. The property breaks as we compute an 
optimal tour {\em passing through} a set of convex polygons, as is 
required by \cvrg. As an example, consider the setup in 
Fig.~\ref{fig:no-dp} with six rectangles and assume that the robot starts 
from $P_1$ and ends at $P_6$. For the visiting sequence 
$P_1P_2P_3P_4P_5P_6$, the blue solid path is the optimal local path. 
On the other hand, for the visiting sequence $P_1P_2P_3P_5P_4P_6$, the 
red dashed path is optimal. We observe that the partial path between 
$P_1$ and $P_3$ cannot be reused. 

Though there is a lack of partial ordering in optimal tour computation 
for \cvrg, for a tour with $k$ sites where $k$ is not large, which is 
the case due to limited robot capacity, sifting through all $k!$ 
sequences is doable provided that the optimal tour based on each sequence can be computed 
efficiently. This turns out to be the case when the regions are convex; 
all that is required is to iteratively improve a tour locally, as outlined 
in Algorithm~\ref{alg:elastic}. The sub-routine \lia attempts to move 
$p_i$ on the polygon $P_i$ to minimize $|p_i-p_{i-1}| + |p_{i+1}-p_{i}|$ based on the current positions of $p_{i-1}$ and $p_{i+1}$. We call 
the algorithm an {\em iterative elastic improvement} as the process 
acts as fixing two endpoints of an elastic band and let a point 
in the middle ``slide'' in a restricted area (e.g. a convex polygon) to a low energy configuration.
\begin{algorithm}
    \SetKwInOut{Input}{Input}
    \SetKwInOut{Output}{Output}
    \SetKwComment{Comment}{\% }{}
    \caption{\eia (Iterative Elastic Improvement)}
		\label{alg:elastic}
    \SetAlgoLined
		\begin{small}
		\vspace{1mm}
    \Input{$P_1, \ldots, P_k$: polygon sequence; d: depot location}
    \Output{$\tau$: the optimal tour, initially empty}
		\vspace{1mm}
    
		\For{each $P_i$, $1\le i \le k$}{
		$p_i \leftarrow $ a random point on $P_i$;\\
		$\tau \leftarrow \tau + p_i$; \Comment{\small{``$+$'' denotes appending}}
		}

		\vspace{1mm}
		$\tau' = \varnothing$;
		\vspace{1mm}

    \While{$\tau \ne \tau'$}{
        $\tau' \leftarrow \tau$; $\tau \leftarrow [];$\\
        \For{$1\le i \le k$}{
						$\tau \leftarrow \tau + $ \lia($i$, $\tau$, $\tau'$, $P_i$, $d$);
				}
    }
		\Return $\tau$;
		\end{small}
\end{algorithm}

To see that Algorithm~\ref{alg:elastic} is globally optimal when the 
polygons $P_1, \ldots, P_k$ are convex, we first examine the case where 
each $P_i$ is a line segment $L_i$. We start with defining the terms 
{\em refraction}, {\em reflection}, {\em crossing}, and 
{\em mirror-reflection}. Without loss of generality, we assume that a 
tour $T$ never overlaps with a line segment $L_i$ as the probability of 
a shortest tour overlapping with a line segment is zero. 

\vspace{-1mm}
\begin{definition}[Refraction, Reflection, Crossing, and 
Mirror-Reflection] Let a line segment $L$ and a tour $T$ intersect at 
$p$. We say $T$ is {\em 
refracted} by $L$ at $p$ if the two line segments $\ell_1$ and $\ell_2$ of 
$T$ meeting at $p$ lie on different sides of $L$. If $\ell_1$ and $\ell_2$ 
are co-linear, then the refraction is a {\em crossing}. Otherwise, if 
$\ell_1$ and $\ell_2$ lie on the same side of $L$, $T$ is {\em reflected} 
by $L$. A reflection is a {\em mirror-reflection} if the bisector of the 
angle between $\ell_1$ and $\ell_2$ is perpendicular to $L$.
\end{definition}
\vspace{-1mm}

\begin{figure}[ht]
\begin{center}
\begin{tikzpicture}
\coordinate (v1) at (-0.1, -0.1);
\coordinate (v2) at (0.9, 0.1);
\coordinate (v3) at (1.6, 0.1);
\coordinate (v4) at (0.1, 0.1);
\draw[line width = 0.4mm,color=black,-] (-1.5,0.75) -- (1.5, 0.75);
\draw[line width = 0.2mm,color=blue,-latex] (-1.0,-0.25) -- (-0.3, 0.75);
\draw[line width = 0.2mm,color=blue,-latex] (-0.3, 0.75) -- (0, 1.55);
\draw[line width = 0.2mm,color=blue,-latex] (-0.5,-0.25) -- (0.0, 0.75);
\draw[line width = 0.2mm,color=blue,-latex] (0.0,0.75) -- (0.8, -0.25);
\node[black] at (-1, 1.2) {\small{refraction}};
\node[black] at (1, 0.5) {\small{reflection}};
\end{tikzpicture}\begin{tikzpicture}
\coordinate (v1) at (-0.1, -0.1);
\coordinate (v2) at (0.9, 0.1);
\coordinate (v3) at (1.6, 0.1);
\coordinate (v4) at (0.1, 0.1);
\draw[line width = 0.4mm,color=black,-] (-1.5,0.75) -- (1.5, 0.75);
\draw[line width = 0.2mm,color=blue,-latex] (-0.9,-0.25) -- (-0.3, 0.75);
\draw[line width = 0.2mm,color=blue,-latex] (-0.3, 0.75) -- (0.3, 1.75);
\draw[line width = 0.2mm,color=blue,-latex] (-0.5,-0.25) -- (0, 0.75);
\draw[line width = 0.2mm,color=blue,-latex] (0,0.75) -- (0.5, -0.25);
\node[black] at (-1, 1.0) {\small{crossing}};
\node[black] at (1.4, 0.5) {\small{mirror-reflection}};
\end{tikzpicture}
\end{center}
\caption{Illustrations of refraction, reflection, crossing, and 
mirror-reflection. The bold horizontal line is a line segment $L$ 
and the line segments with arrows are part of a tour $T$.}
\label{fig:rrcm}
\vspace{-1mm}
\end{figure}
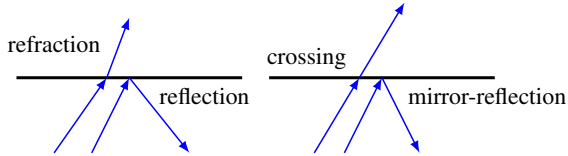

These definitions are illustrated in Fig.~\ref{fig:rrcm}. 

\vspace{-1mm}
\begin{definition}[Elastic Band]
Given a point $d \in R^2$ and a set of line segments $L_1, \ldots, L_k$, 
we say a tour $T$ starting at $d$ and passing through $L_1, \ldots, L_k$ 
in that order, at locations $p_1, \ldots, p_k$, is an elastic band if for 
any $i$, $1 \le i \le k$, one of the following holds:
1) $T$ is mirror-reflected by $L_i$ at $p_i$, 2) $T$ crosses $L_i$ at $p_i$, 
or 3) $p_i$ is an end point of $L_i$ and the angle formed by the two 
line segment of $T$ meeting at $p_i$ and enclosing $L_i$ is no less 
than $\pi$. 
\end{definition}
\vspace{-1mm}

The conditions specified in the definition of elastic bands are necessary
conditions for a tour to be locally optimal (shortest).
We note that a tour $T$ may intersect an $L_i$ at more than one point. 
However, since the visiting order of $L_1, \ldots, L_k$ is fixed, 
$p_i$'s are uniquely defined. It is clear that a shortest tour $T$ going 
through $L_1, \ldots, L_k$, in that order, must be an elastic band: if at 
some $p_i$ the elastic band properties are not satisfied, $T$ can be 
readily shortened. 

\vspace{-1mm}
\begin{lemma}
A shortest tour $T$ starting at a depot $d$ and going through line 
segments $L_1, \ldots, L_k$, in that order, at points $p_1, \ldots, 
p_k$, must be an elastic band. 
\end{lemma}
\vspace{-1mm}

Next, we will show uniqueness of elastic bands, which yields
global optimality. The main idea behind the proof of uniqueness
is to establish that two different optimal elastic bands going
through the same set of line segments cannot meet back at the 
starting point (depot $d$). Showing this requires detailed 
cases analysis. We start with some necessary 
terminologies, definitions, and intermediate lemmas. 

Let $T$ be a tour intersecting line segment $L_i$ 
at $p$. As shown in Fig.~\ref{fig:properties}, $pp'$ 
is part of $L_i$, where $p$ may be an endpoint of $L_i$. 
If $T$ is an elastic band, then we have 
$\theta_1\leq \theta_2$ in the setup given in Fig. 
\ref{fig:bothreflectiona} and $\theta_3 \leq \theta_4$ in the 
setup given in Fig. \ref{fig:bothreflectionb}. Otherwise, $p$ 
can be moved toward the middle of $pp'$ to reduce the length 
of $T$. 
Specifically, $\theta_1<\theta_2$ and $\theta_3<\theta_4$ only 
when $p$ is an endpoint of $L_i$.

\usetikzlibrary{quotes,angles}
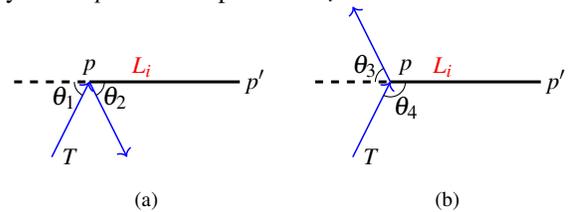
\begin{figure}[ht]
\begin{center}
\begin{subfigure}[b]{0.2\textwidth}
\begin{tikzpicture}
\coordinate (p1) at (-1.0, 0.75);
\coordinate (e1) at (-2.0, 0.75);
\coordinate (e2) at (1.0, 0.75);
\coordinate (T11) at (-1.5,-0.25);
\coordinate (T12) at (-0.5, -0.25);

\draw[line width = 0.4mm, dashed, color=black,-] (e1) -- (p1);
\draw[line width = 0.4mm, color=black,-] (p1) -- (e2);

\draw[line width = 0.2mm,color=blue,->] (T11) -- (p1);
\draw[line width = 0.2mm,color=blue,->] (p1) -- (T12);

\node[black] at (-1.25, -0.25) {\small{$T$}};
\node[black] at (1.2, 0.75) {\small{$p'$}};
\node[red] at (-0.3, 0.95) {\small{$L_i$}};
\node[black] at (-1.0, 0.95) {\small{$p$}};

\pic [draw, -, "$\theta_1$", angle eccentricity=2.0,angle radius=2mm] {angle = e1--p1--T11};
\pic [draw, -, "$\theta_2$", angle eccentricity=2.0,angle radius=2mm] {angle = T12--p1--e2};

\end{tikzpicture}
\caption{}
\label{fig:bothreflectiona}
\end{subfigure}
\hspace{2mm}
\begin{subfigure}[b]{0.2\textwidth}
\begin{tikzpicture}
\coordinate (p1) at (-1.0, 0.75);
\coordinate (e1) at (-2.0, 0.75);
\coordinate (e2) at (1.0, 0.75);
\coordinate (T11) at (-1.5,-0.25);
\coordinate (T12) at (-1.5, 1.75);

\draw[line width = 0.4mm,dashed, color=black,-] (e1) -- (p1);
\draw[line width = 0.4mm, color=black,-] (p1) -- (e2);

\draw[line width = 0.2mm,color=blue,->] (T11) -- (p1);
\draw[line width = 0.2mm,color=blue,->] (p1) -- (T12);

\node[black] at (-1.25, -0.25) {\small{$T$}};
\node[black] at (1.2, 0.75) {\small{$p'$}};
\node[red] at (-0.3, 0.95) {\small{$L_i$}};
\node[black] at (-0.8, 0.95) {\small{$p$}};

\pic [draw, -, "$\theta_3$", angle eccentricity=2.0,angle radius=2mm] {angle = T12--p1--e1};
\pic [draw, -, "$\theta_4$", angle eccentricity=2.0,angle radius=2mm] {angle = T11--p1--e2};

\end{tikzpicture}
\caption{}
\label{fig:bothreflectionb}
\end{subfigure}
\end{center}
\vspace{-3mm}
\caption{ Given an elastic band $T$ and the solid line is part of the line segment $L_i$, we have $\theta_1\leq\theta_2$, $\theta_3\leq\theta_4$.}
\label{fig:properties}
\vspace{-1mm}
\end{figure}

Given an elastic band $T_j$ starting at $d$ and intersecting with line 
segments $L_1, \ldots, L_k$ in the given order at $p_{j,1}, \ldots, 
p_{j,k}$, respectively, we define a {\em ray} $R_{j,i}$ as follows.
$$R_{j,i}=
\begin{cases}\text{ray }\overrightarrow{dp_{j,1}}, & \text{for } i=0\\
\text{ray }\overrightarrow{p_{j,i}p_{j,i+1}}, & \text{for } 0<i<k\\
\text{ray }\overrightarrow{p_{j,k}d}, & \text{for } i=k
\end{cases}$$
For discussing the relationship between two elastic bands, we 
introduce a definition {\em convergence angle} $v$ for rays.
\begin{definition}
Let $R_1$ and $R_2$ be two rays with direction vectors 
$\overrightarrow{n_1}$ and $\overrightarrow{n_2}$, respectively. 
If $R_1$ and $R_2$ intersect, then the {\em convergence angle} $v$ is 
defined to be the angle between $\overrightarrow{n_1}$ and 
$\overrightarrow{n_2}$, i.e. $\arccos(\overrightarrow{n_1}, 
\overrightarrow{n_2})$, as $\theta_1$ in Fig. 
\ref{fig:convergence-angle-1}. 
On the other hand, if the two rays do not intersect, then the 
convergence angle $v$ is defined as $-\arccos(\overrightarrow{n_1}, 
\overrightarrow{n_2})$ (e.g., $-\theta_2$ in Fig. 
\ref{fig:convergence-angle-2}). 
\end{definition}

We note that the later case in the definition of the convergence 
angle includes two sub cases: two rays intersecting on their 
extensions (along the negative directions of the rays) or one ray 
intersecting the extension of another ray. 
According to the definition, two different rays intersect if and only if they have a positive convergence angle.

\usetikzlibrary{quotes,angles}
\begin{figure}[ht]
\begin{center}
\begin{subfigure}[b]{0.35\columnwidth}
\begin{center}
\begin{tikzpicture}
\coordinate (p1) at (-1.0, 0.75);
\coordinate (p2) at (0.8, 0.75);
\coordinate (p3) at (-0.4, 2.75);
\coordinate (p4) at (-0.25, 3.25);
\coordinate (p5) at (-0.7, 3.25);


\draw[line width = 0.2mm,color=blue,-] (p1) -- (p3);
\draw[line width = 0.2mm,color=blue,-] (p2) -- (p3);
\draw[line width = 0.2mm,color=blue,->] (p3) -- (p4);
\draw[line width = 0.2mm,color=blue,->] (p3) -- (p5);

\node[black] at (-1.3, 1.25) {\small{$R_1$}};
\node[black] at (1.0, 1.25) {\small{$R_2$}};


\pic [draw, -, "$\theta_1$", angle eccentricity=2.0,angle radius=2mm] {angle = p1--p3--p2};

\end{tikzpicture}
\caption{\label{fig:convergence-angle-1}}
\end{center}
\end{subfigure}
\begin{subfigure}[b]{0.35\columnwidth}
\begin{center}
\begin{tikzpicture}
\coordinate (p1) at (-1.0, 1.75);
\coordinate (p2) at (0.8, 1.75);
\coordinate (p3) at (-0.4, -0.75);
\coordinate (m1) at (-0.7, 0.5);
\coordinate (m2) at (0.2, 0.5);

\draw[line width = 0.2mm, dashed,color=blue,->] (p3) -- (p1);
\draw[line width = 0.2mm, dashed, color=blue,->] (p3) -- (p2);
\draw[line width = 0.2mm, color=blue,->] (m1) -- (p1);
\draw[line width = 0.2mm, color=blue,->] (m2) -- (p2);

\node[black] at (-1.0, 0.25) {\small{$R_1$}};
\node[black] at (0.5, 0.25) {\small{$R_2$}};

\pic [draw, -, "$\theta_2$", angle eccentricity=3.0,angle radius=2mm] {angle = p2--p3--p1};

\end{tikzpicture}
\caption{\label{fig:convergence-angle-2}\hspace{30mm}}
\end{center}
\end{subfigure}
\end{center}
\vspace{-2mm}
\caption{\label{fig:ca}Two possible arrangements of two rays. 
In the second case, the intersection may be between one ray and the 
extension of a second ray.}
\vspace{-1mm}
\end{figure}
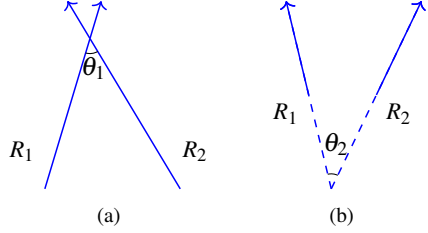

\vspace{-1mm}
\begin{definition}
Let $T_1$ and $T_2$ be two elastic bands going through a line segment 
$L_i$ containing rays $\overrightarrow{p_{1,i}p_{1,i+1}}$ and 
$\overrightarrow{p_{2,i}p_{2,i+1}}$, respectively. When the two 
outgoing rays emit from  $L_i$ from the opposite sides and it holds 
that
\begin{equation}
\label{ieq:zigzag}
\angle p_{1,i+1}p_{1,i}p_{2,i} + \angle p_{2,i+1}p_{2,i}p_{1,i} < \pi,
\end{equation}
\vspace{-1mm}
we say that the rays form a {\em zigzag} with respect to $L_i$.
\end{definition}
\vspace{-1mm}

\begin{figure}[ht]
\centering
\begin{tikzpicture}
\coordinate (p1) at (-1.0, 0.75);
\coordinate (p2) at (0.8, 0.75);
\coordinate (p3) at (-0.5, 1.45);
\coordinate (p4) at (1.2, 0.05);
\coordinate (e1) at (-1.5, 0.75);
\coordinate (e2) at (1.3, 0.75);

\draw[line width = 0.4mm,dashed,color=black,-] (e1) -- (e2);
\draw[line width = 0.4mm,color=black,-] (p1) -- (p2);

\draw[line width = 0.2mm,color=blue,->] (p1) -- (p3);
\draw[line width = 0.2mm,color=blue,->] (p2) -- (p4);

\node[black] at (-1.3, 1.25) {\small{$T_1$}};
\node[black] at (1.3, 0.45) {\small{$T_2$}};
\node[black] at (-1.0, 0.55) {\small{$p_{1,i}$}};
\node[black] at (-0.3, 1.65) {\small{$p_{1,i+1}$}};
\node[black] at (0.8, 0.95) {\small{$p_{2,i}$}};
\node[black] at (1.2, -0.15) {\small{$p_{2,i+1}$}};
\node[red] at (-0.1, 0.95) {\small{$L_i$}};

\pic [draw, -, "", angle eccentricity=2.0,angle radius=2mm] {angle = p2--p1--p3};

\pic [draw, -, "", angle eccentricity=2.0,angle radius=2mm] {angle = p1--p2--p4};
\end{tikzpicture}
\caption{ A zigzag is the case when the two tours 
leaving a line segment from the opposite sides and 
$\angle p_{1,i+1}p_{1,i}p_{2,i} + \angle p_{2,i+1}p_{2,i}p_{1,i} < \pi$.}
\label{fig:zigzag}
\end{figure}
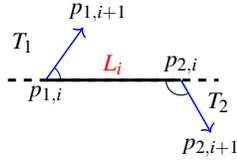

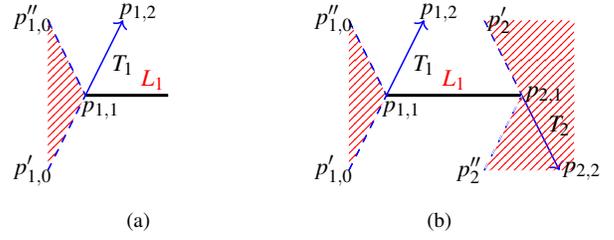
\begin{figure}[ht]
\begin{center}
\begin{subfigure}[b]{0.2\textwidth}
\begin{tikzpicture}
\coordinate (p1) at (-1.0, 0.75);
\coordinate (p2) at (0.1, 0.75);
\coordinate (e1) at (-2.0, 0.75);
\coordinate (e2) at (1.8, 0.75);
\coordinate (T11) at (-1.5,-0.25);
\coordinate (T111) at (-1.5,1.75);
\coordinate (T12) at (-0.5, 1.75);
\coordinate (v4) at (0.1, 0.1);

\draw[line width = 0.4mm,color=black,-] (p1) -- (p2);

\draw[line width = 0.2mm,dashed,color=blue,-] (T11) -- (p1);
\draw[line width = 0.2mm,dashed,color=blue,-] (T111) -- (p1);
\draw[line width = 0.2mm,color=blue,->] (p1) -- (T12);

\node[black] at (-0.5, 1.15) {\small{$T_1$}};
\node[black] at (-0.8, 0.55) {\small{$p_{1,1}$}};
\node[black] at (-0.3, 1.85) {\small{$p_{1,2}$}};
\node[black] at (-1.7,-0.25) {\small{$p_{1,0}'$}};
\node[black] at (-1.7,1.75) {\small{$p_{1,0}''$}};
\node[red] at (-0.1, 0.95) {\small{$L_1$}};
\draw[dashed, blue,pattern=north east lines,pattern color=red] (T11) -- (p1)--(T111);


\end{tikzpicture}
\caption{}
\label{fig:first-obj-proof1}
\end{subfigure}
\hspace{2mm}
\begin{subfigure}[b]{0.2\textwidth}
\begin{tikzpicture}
\coordinate (p1) at (-1.0, 0.75);
\coordinate (p2) at (0.8, 0.75);
\coordinate (e1) at (-2.0, 0.75);
\coordinate (e2) at (1.8, 0.75);
\coordinate (T11) at (-1.5,-0.25);
\coordinate (T111) at (-1.5,1.75);
\coordinate (T12) at (-0.5, 1.75);
\coordinate (T21) at (0.3,-0.25);
\coordinate (T211) at (0.3,1.75);
\coordinate (T22) at (1.3, -0.25);
\coordinate (v4) at (0.1, 0.1);

\draw[line width = 0.4mm,color=black,-] (p1) -- (p2);

\draw[line width = 0.2mm,dashed,color=blue,-] (T11) -- (p1);
\draw[line width = 0.2mm,dashed,color=blue,-] (T111) -- (p1);
\draw[line width = 0.2mm,color=blue,->] (p1) -- (T12);

\draw[line width = 0.2mm, dashed,color=blue,-] (T21) -- (p2);
\draw[line width = 0.2mm, dashed,color=blue,-] (T211) -- (p2);
\draw[line width = 0.2mm,color=blue,->] (p2) -- (T22);
\draw[dashed, blue,pattern=north east lines,pattern color=red] (T11) -- (p1)--(T111);
\draw[dashed, white,pattern=north east lines,pattern color=red] (T211) -- (1.5, 1.75) -- (1.5, -0.25)--(T21) -- (p2);
\node[black] at (-0.5, 1.15) {\small{$T_1$}};
\node[black] at (-0.3, 1.85) {\small{$p_{1,2}$}};
\node[black] at (-0.8, 0.55) {\small{$p_{1,1}$}};
\node[black] at (1.1, 0.75) {\small{$p_{2,1}$}};
\node[black] at (-1.7,-0.25) {\small{$p_{1,0}'$}};
\node[black] at (-1.7,1.75) {\small{$p_{1,0}''$}};
\node[black] at (0.1,-0.25) {\small{$p_{2}''$}};
\node[black] at (0.5,1.75) {\small{$p_{2}'$}};
\node[black] at (1.3, 0.35) {\small{$T_{2}$}};
\node[black] at (1.6, -0.25) {\small{$p_{2,2}$}};
\node[red] at (-0.1, 0.95) {\small{$L_1$}};



\end{tikzpicture}
\caption{}
\label{fig:first-obj-proof2}
\end{subfigure}
\end{center}
\vspace{-2mm}
\caption{\footnotesize If $T_1$ and $T_2$ go through $L_1$ with a zigzag, then there is no proper location for the depot $d$.}
\label{fig:first-obj-proofs}
\vspace{-1mm}
\end{figure}

\vspace{-1mm}
\begin{lemma}\label{lm:first-object}
Given two elastic bands $T_1$ and $T_2$ going through line segment $L_1$ containing rays $\overrightarrow{p_{1,1}p_{1,2}}$ and 
$\overrightarrow{p_{2,1}p_{2,2}}$, respectively. If $L_1$ is the first line segment in the visiting order, then the rays $\overrightarrow{p_{1,1}p_{1,2}}$ and
$\overrightarrow{p_{2,1}p_{2,2}}$ cannot form a zigzag with respect to $L_1$.
\end{lemma}
\vspace{-3mm}
\begin{proof}
Since $L_1$ is the first object in the visiting order, both $T_1$ and $T_2$ 
come from the same point $d$ before going through $L_1$. We prove the lemma 
by showing that $d$ cannot exist 
if $\overrightarrow{p_{1,1}p_{1,2}}$ and
$\overrightarrow{p_{2,1}p_{2,2}}$ form a zigzag with respect to $L_1$.

Denote the intersection of elastic band $T_i$ and line segment $L_j$ by $p_{i,j}$. 
If a zigzag forms about $L_1$, then the inequality (\ref{ieq:zigzag}) holds, 
i.e. $\angle p_{1,2}p_{1,1}p_{2,1} + \angle p_{2,2}p_{2,1}p_{1,1}<\pi$. Without 
loss of generality, we may assume $\angle p_{1,2}p_{1,1}p_{2,1} < \pi/2$. Due to 
the properties of elastic bands, $d$ must be in the shaded area as shown in 
Fig. \ref{fig:first-obj-proof1}, where $p_{1,0}'p_{1,1}p_{1,2}$ is a crossing 
and $p_{1,0}''p_{1,1}p_{1,2}$ is a mirror-reflection. Once 
$\angle p_{2,2}p_{2,1}p_{1,1}$ is fixed, there will be another shaded area 
corresponding to $T_2$, and $d$ has to be in the intersection of the two areas. 
To allow overlapping between the areas, $\angle p_{2,2}p_{2,1}p_{1,1}$ must 
be larger than $\pi/2$ and the dash area is as shown in Fig. 
\ref{fig:first-obj-proof2}. With the inequality~\eqref{ieq:zigzag}, the 
aforementioned two shaded areas must be disjoint, which leads to the nonexistence 
of $d$.
\end{proof}

\vspace{-1mm}
\begin{lemma}\label{lm:impossible-zigzag}
Consider two elastic bands $T_1$ and $T_2$, $v_i$ is the convergence angle of the corresponding rays $R_{1,i}$ and $R_{2,i}$. It is impossible to have a zigzag for an object $L_j$ if $v_i\in (-\pi, 0]$ holds for all $i = 1,2,..., j-1$.
\end{lemma}
\vspace{-1mm}
\begin{proof}
Assuming the contrary, then~\eqref{ieq:zigzag} holds. Without 
loss of generality, we can assume $\angle p_{1,j+1}p_{1,j}p_{2,j} < \pi/2$.

\begin{enumerate}
\item When $\angle p_{2,j+1}p_{2,j}p_{1,j} < \pi/2$, by the assumption 
$-\pi < v_{j-1}\leq 0$, it is easy to verify that the tours must go through 
object $L_{j-1}$ with a zigzag and $\angle p_{1,j}p_{1,j-1}p_{2,j-1}$, 
$\angle p_{2,j}p_{2,j-1}p_{1,j-1}<\pi/2$. With the assumption that 
$v_i\in (-\pi, 0]$ always holds for all $i = 1,2,..., j-1$, recursively, 
the two tours must leave the first object with a zigzag,
contradicting Lemma \ref{lm:first-object}.
    
\item When $\angle p_{2,j+1}p_{2,j}p_{1,j} \geq \pi/2$, there are 
four cases: a) Both $T_1$ and $T_2$ pass $L_j$ with refractions,
        b) Both $T_1$ and $T_2$ pass $L_j$ with reflections,
        c) $T_1$ passes $L_j$ with a reflection and $T_2$ passes $L_j$ with a refraction, or
        d) $T_1$ passes $L_j$ with a refraction and $T_2$ passes $L_j$ with a reflection. 
    For the first case, since $ v_{j-1} \leq 0$, $L_{j-1}$ intersects $L_j$ as 
		shown in Fig.~\ref{fig:zigzag-proof} where $\theta_3 < \theta_1, \theta_4 
		< \theta_2$, we must have $\theta_3 + \theta_4 <\pi$. Therefore, the tours 
		emanate from object $L_{j-1}$ with a zigzag. Recursively, it leads to a 
		contradiction to the statement of  Lemma \ref{lm:first-object}. The other cases 
		can be verified with similar reasoning.
\end{enumerate}
\end{proof}		
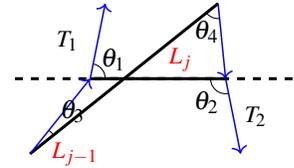
\begin{figure}[ht]
\centering
\begin{tikzpicture}
\coordinate (p1) at (-1.0, 0.75);
\coordinate (p2) at (0.8, 0.75);
\coordinate (e1) at (-2.0, 0.75);
\coordinate (e2) at (1.8, 0.75);
\coordinate (p3) at (-0.8, 1.75);
\coordinate (p4) at (1.0, -0.25);
\coordinate (p5) at (-1.8, -0.25);
\coordinate (p6) at (0.7, 1.75);

\draw[line width = 0.4mm,color=black,-] (p1) -- (p2);
\draw[line width = 0.4mm,color=black,dashed,-] (e1) -- (e2);
\draw[line width = 0.4mm,color=black,-] (p5) -- (p6);

\draw[line width = 0.2mm,color=blue,->] (p1) -- (p3);
\draw[line width = 0.2mm,color=blue,->] (p2) -- (p4);
\draw[line width = 0.2mm,color=blue,->] (p5) -- (p1);
\draw[line width = 0.2mm,color=blue,->] (p6) -- (p2);

\node[black] at (-1.3, 1.25) {\small{$T_1$}};
\node[black] at (1.2, 0.25) {\small{$T_2$}};
\node[red] at (0.2, 1) {\small{$L_j$}};
\node[red] at (-1.2, -0.25) {\small{$L_{j-1}$}};

\pic [draw, -, "$\theta_1$", angle eccentricity=2.0,angle radius=2mm] {angle = p2--p1--p3};
\pic [draw, -, "$\theta_2$", angle eccentricity=2.0,angle radius=2mm] {angle = p1--p2--p4};
\pic [draw, -, "$\theta_4$", angle eccentricity=2.0,angle radius=2mm] {angle = p5--p6--p2};
\pic [draw, -, "$\theta_3$", angle eccentricity=2.0,angle radius=4mm] {angle = p6--p5--p1};

\end{tikzpicture}
\caption{ When $\angle p_{2,j+1}p_{2,j}p_{1,j} \geq \pi/2$, a contradiction
arises.}
		\label{fig:zigzag-proof}
\vspace{-2mm}
\end{figure}


\vspace{-1mm}
\begin{theorem}\label{thm:uniqueness}
For a tour starting from the depot and passing through a set of straight
line segments in a fixed order, its length has a unique global minimum as 
realized by an elastic band. 
\end{theorem}
\vspace{-3mm}
\begin{proof}
Assuming the contrary, then we can let $T_1$ and $T_2$ be two different 
elastic bands for the problem. Recall that $v_i$ is the convergence angle 
of rays $\overrightarrow{p_{1,i}p_{1,i+1}}$ and 
$\overrightarrow{p_{2,i}p_{2,i+1}}$, where $p_{t,i}$ is the intersection 
between tour $t$ and line segment $L_i$. Based on the fact that both 
tours share the same depot, $T_1$ and $T_2$ must diverge at some point 
and converge before the destination, i.e. $\exists (i,j)$, s.t. $i<j$, 
$v_{i}<0<v_{j}$. We show it is impossible to have the first $j$ of this 
kind. By definition, the range of convergence angle is $[-\pi, \pi]$. 
Since the probability of tours overlapping with a line segment 
object is zero, we may assume that $v_i, v_j \in ( -\pi, \pi)$.

There are three cases which may allow $v_j>0$:
C--1: $T_1$ and $T_2$ are from the same side of $L_j$ and pass 
with refractions;
C--2: $T_1$ and $T_2$ are from the same side of $L_j$ and pass 
		with reflections;
and C--3: $T_1$ and $T_2$ are from different sides of $L_j$ and with 
		different types of passing.

For case C--1 (as Fig. \ref{fig:same-side-refraction}), $v_{j-1} = -(\pi - \theta_2 - \theta_4) = - \pi + \theta_2 + \theta_4 \leq 0 $ and $v_{j} = \pi - (\pi - \theta_1) - (\pi - \theta_3)= - \pi + \theta_1 + \theta_3$.
With the properties of elastic bands, we have $0<\theta_1 \leq \theta_2$, $0<\theta_3 \leq \theta_4$, therefore $-\pi < v_{j}\leq v_{j-1} \leq 0$.

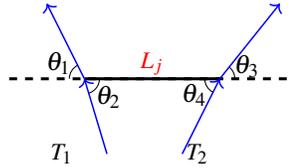
\begin{figure}[ht]
    \centering
    \begin{tikzpicture}
\coordinate (p1) at (-1.0, 0.75);
\coordinate (p2) at (0.8, 0.75);
\coordinate (e1) at (-2.0, 0.75);
\coordinate (e2) at (1.8, 0.75);
\coordinate (T11) at (-0.7,-0.25);
\coordinate (T12) at (-1.5, 1.75);
\coordinate (T21) at (0.3,-0.25);
\coordinate (T22) at (1.6, 1.75);

\draw[line width = 0.4mm,color=black,-] (p1) -- (p2);
\draw[line width = 0.4mm,dashed, color=black,-] (e1) -- (e2);

\draw[line width = 0.2mm,color=blue,->] (T11) -- (p1);
\draw[line width = 0.2mm,color=blue,->] (p1) -- (T12);

\draw[line width = 0.2mm,color=blue,->] (T21) -- (p2);
\draw[line width = 0.2mm,color=blue,->] (p2) -- (T22);

\node[black] at (-1.3, -0.25) {\small{$T_1$}};
\node[black] at (0.5, -0.25) {\small{$T_2$}};
\node[red] at (-0.1, 0.95) {\small{$L_j$}};

\pic [draw, -, "$\theta_2$", angle eccentricity=2.0,angle radius=2mm] {angle = T11--p1--e2};
\pic [draw, -, "$\theta_1$", angle eccentricity=2.0,angle radius=2mm] {angle = T12--p1--e1};

\pic [draw, -, "$\theta_3$", angle eccentricity=2.0,angle radius=2mm] {angle = e2--p2--T22};
\pic [draw, -, "$\theta_4$", angle eccentricity=2.0,angle radius=2mm] {angle = e1--p2--T21};
\end{tikzpicture}
\caption{When $-\pi<v_{j-1}\leq 0$, if two elastic bands $T_1$ and $T_2$ 
approach object $L_j$ from the same side and form refractions, then
$-\pi<v_j\leq 0$.}
\label{fig:same-side-refraction}
\end{figure}
    
With similar reasoning, we can verify that $-\pi<v_{j}\leq 0$, when $-\pi<v_{j-1}\leq 0$ 
in case C--2.
    
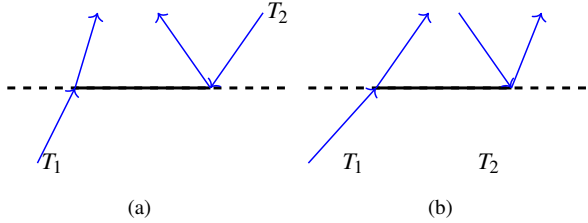
\begin{figure}[ht]
\begin{center}
\begin{subfigure}[b]{0.2\textwidth}
\begin{tikzpicture}
\coordinate (p1) at (-1.0, 0.75);
\coordinate (p2) at (0.8, 0.75);
\coordinate (e1) at (-1.9, 0.75);
\coordinate (e2) at (1.8, 0.75);
\coordinate (T11) at (-1.5,-0.25);
\coordinate (T12) at (-0.7, 1.75);
\coordinate (T21) at (1.5,1.75);
\coordinate (T22) at (0.1, 1.75);

\draw[line width = 0.4mm,color=black,-] (p1) -- (p2);
\draw[line width = 0.4mm,dashed, color=black,-] (e1) -- (e2);

\draw[line width = 0.2mm,color=blue,->] (T11) -- (p1);
\draw[line width = 0.2mm,color=blue,->] (p1) -- (T12);

\draw[line width = 0.2mm,color=blue,->] (T21) -- (p2);
\draw[line width = 0.2mm,color=blue,->] (p2) -- (T22);
\node[black] at (-1.3, -0.25) {\small{$T_1$}};
\node[black] at (1.7, 1.75) {\small{$T_2$}};



\end{tikzpicture}
\caption{}
\label{fig:zigzag-case1}
\end{subfigure}
\hspace{2mm}
\begin{subfigure}[b]{0.2\textwidth}
\begin{tikzpicture}
\coordinate (p1) at (-1.0, 0.75);
\coordinate (p2) at (0.8, 0.75);
\coordinate (e1) at (-1.9, 0.75);
\coordinate (e2) at (1.8, 0.75);
\coordinate (T11) at (-1.9,-0.25);
\coordinate (T12) at (-0.3, 1.75);
\coordinate (T21) at (0.1,1.75);
\coordinate (T22) at (1.2, 1.75);

\draw[line width = 0.4mm,color=black,-] (p1) -- (p2);
\draw[line width = 0.4mm,dashed, color=black,-] (e1) -- (e2);

\draw[line width = 0.2mm,color=blue,->] (T11) -- (p1);
\draw[line width = 0.2mm,color=blue,->] (p1) -- (T12);

\draw[line width = 0.2mm,color=blue,->] (T21) -- (p2);
\draw[line width = 0.2mm,color=blue,->] (p2) -- (T22);
\node[black] at (-1.3, -0.25) {\small{$T_1$}};
\node[black] at (0.5, -0.25) {\small{$T_2$}};


\end{tikzpicture}
\caption{}
\label{fig:zigzag-case2}
\end{subfigure}
\end{center}
\vspace{-2mm}
\caption{Two sub-cases in case C--3.}
\vspace{-1mm}
\label{fig:zigzag-cases}
\end{figure}

As for case C--3, there are two sub-cases (Fig.~\ref{fig:zigzag-case1} 
and Fig.~\ref{fig:zigzag-case2}). For the first sub-case, assuming that $-\pi 
<v_i\leq 0$, for all $i < j$, $\overrightarrow{p_{1,j-1}p_{1,j}}$ and 
$\overrightarrow{p_{2,j-1}p_{2,j}}$ must be from the opposite sides 
of object $L_{j-1}$ as shown in Fig. \ref{fig:zigzag-case1-proof}.

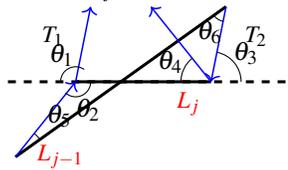
\begin{figure}[ht]
    \centering
\begin{tikzpicture}
\coordinate (p1) at (-1.0, 0.75);
\coordinate (p2) at (0.8, 0.75);
\coordinate (e1) at (-1.9, 0.75);
\coordinate (e2) at (1.8, 0.75);
\coordinate (p3) at (-0.8, 1.75);
\coordinate (p4) at (0.0, 1.75);
\coordinate (p5) at (-1.8, -0.25);
\coordinate (p6) at (1.0, 1.75);

\draw[line width = 0.4mm,color=black,-] (p1) -- (p2);
\draw[line width = 0.4mm,color=black,dashed,-] (e1) -- (e2);
\draw[line width = 0.4mm,color=black,-] (p5) -- (p6);

\draw[line width = 0.2mm,color=blue,->] (p1) -- (p3);
\draw[line width = 0.2mm,color=blue,->] (p2) -- (p4);
\draw[line width = 0.2mm,color=blue,->] (p5) -- (p1);
\draw[line width = 0.2mm,color=blue,->] (p6) -- (p2);

\node[black] at (-1.3, 1.35) {\small{$T_1$}};
\node[black] at (1.4, 1.35) {\small{$T_2$}};
\node[red] at (0.5, 0.45) {\small{$L_j$}};
\node[red] at (-1.2, -0.25) {\small{$L_{j-1}$}};

\pic [draw, -, "$\theta_1$", angle eccentricity=2.0,angle radius=2mm] {angle = T12--p1--e1};
\pic [draw, -, "$\theta_2$", angle eccentricity=2,angle radius=2mm] {angle = T11--p1--p2};
\pic [draw, -, "$\theta_3$", angle eccentricity=1.5,angle radius=4mm] {angle = e2--p2--p6};
\pic [draw, -, "$\theta_4$", angle eccentricity=1.5,angle radius=4mm] {angle = p4--p2--p1};
\pic [draw, -, "$\theta_5$", angle eccentricity=2.0,angle radius=4mm] {angle = p6--p5--p1};
\pic [draw, -, "$\theta_6$", angle eccentricity=2.0,angle radius=2mm] {angle = p5--p6--p2};
\end{tikzpicture}
\caption{Geometric analysis shows 
that $-\pi < v_j < 0$ in case C--3.}
\label{fig:zigzag-case1-proof}
\end{figure}

To allow $v_{j}>0$, $\theta_4 < \theta_1$. By the properties 
of the elastic band, $\theta_1 \leq \theta_2, \theta_3 \leq \theta_4$. 
Therefore, $\theta_6 < \theta_3 \leq \theta_4 < \theta_1 \leq \theta_2$. 
Since $\theta_5 + \theta_6 < \theta_5 + \theta_2 < \pi$, the object 
$L_{j-1}$ is gone through with a zigzag, which is impossible by 
Lemma \ref{lm:impossible-zigzag}.

The same conclusion holds for the second sub-case under case C--3. 
Therefore, none of the three cases (C--1, C--2, C--3) allows 
$v_j>0$. And we conclude that once $T_1$ and $T_2$ diverge from one 
point, they are unable to converge to the same destination.
\end{proof}
\vspace{-1mm}

When the visiting sequence consists of line segments, 
Theorem~\ref{thm:uniqueness} ensures the uniqueness of the elastic 
band for each instance. Since a shortest tour is an elastic band, 
the solution we get from Algorithm~\ref{alg:elastic} is optimal. 
Optimality of Algorithm~\ref{alg:elastic} extends when
objects are convex.
\vspace{-1mm}
\begin{corollary}\label{coro:convex}
For a tour starting from the depot passing through a set of convex 
regions in a fixed order, its length has a unique global minimum,
realized by an elastic band. 
\end{corollary}
\vspace{-3mm}
\begin{proof}
Assuming the contrary, we can let $T_1$ and $T_2$ be two different 
elastic bands. Denote the intersection of elastic band $T_i$ and 
convex region $R_j$ by $p_{i,j}$. Recall that the proof of 
Theorem~\ref{thm:uniqueness} is based on the fact that segment 
$p_{1,j}p_{2,j}$ is part of line segment $L_j$. For each convex 
region $R_j$, segment $p_{1,j}p_{2,j}$ is still part of it. 
Therefore, the conclusion holds for the convex case, i.e., 
elastic bands $T_1$ and $T_2$ should be the same tour.
\end{proof}
\vspace{-1mm}

\begin{figure}[h!]
\vspace{1mm}
    \centering
    \begin{tikzpicture}
    \coordinate (p11) at (-0.25, -0.5);
    \coordinate (p12) at (0.0, 0.0);
    \coordinate (p13) at (0.25, -0.5);
    \coordinate (p14) at (0.0, 0.50);
    \coordinate (p21) at (-1.0, 0.00);
    \coordinate (p22) at (0.0, 1.2);
    \coordinate (p221) at (-1.0, 0.9);
    \coordinate (p223) at (1.0, 0.9);
    \coordinate (p23) at (1.0, 0.00);
    \coordinate (p24) at (0.0, 2.0);
    \coordinate (p241) at (-1.5, 1.5);
    \coordinate (p243) at (1.5, 1.5);
    \coordinate (d) at (0.0,- 1.0);
    
    \draw { (p11) -- (p12) -- (p13) -- (p14) } [fill=gray];
    \draw { (p21) -- (p221) -- (p22) -- (p223) -- (p23) -- (p243) -- (p24) -- (p241) } [fill=gray];
    \draw[line width = 0.2mm,color=blue,->] (d) -- (p11);
    \draw[line width = 0.2mm,color=blue,->] (p11) -- (p21);
    \draw[line width = 0.2mm,color=blue,->] (p21) -- (d);
    \draw[line width = 0.2mm,color=blue,->] (d) -- (p13);
    \draw[line width = 0.2mm,color=blue,->] (p13) -- (p23);
    \draw[line width = 0.2mm,color=blue,->] (p23) -- (d);
    
    \node[black] at (0.2, -1) {\small{$d$}}; 
    \node[black] at (0.3, 0.50) {\small{$R_1$}};
    \node[black] at (1.8, 1.5) {\small{$R_2$}};
    \node[black] at (-1.5, 0.0) {\small{$T_1$}};
    \node[black] at (1.5, 0.0) {\small{$T_2$}};
    
    \end{tikzpicture}
    \caption{\footnotesize An instance that has two different elastic bands 
		when some of the regions are non-convex. $R_1$, $R_2$ are two non-convex 
		regions and $d$ is the depot. When the visiting sequence is fixed as 
		$P_1P_2$, $T_1$ and $T_2$ are two tours represented with blue arrows. 
		Both of the tours are elastic bands.}
    \label{fig:counterExample}
		\vspace{-2mm}
\end{figure}
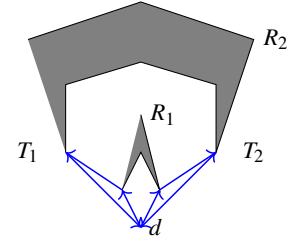

The uniqueness property stated in Corollary~\ref{coro:convex} breaks 
down without the convexity assumption. Fig.~\ref{fig:counterExample} 
shows such an instance with two different elastic bands.

\vspace{-1mm}
\subsection{Handling Overlaps}\label{subsec:overlap}
\vspace{-1mm}
Polygons in \cvrg may overlap (e.g., in a clutter removal scenario). For 
arbitrary polygonal objects $i$ and $j$, if $i$ is placed on $j$ or the 
opposite, we say object $i$ and $j$ overlap with each other. When 
overlaps occur, it is necessary to consider the partial orders before 
we schedule the pickup sequence. When the objects form a stack, we 
should always first pick up the objects on the top. We developed the 
corresponding DP
algorithm for the overlapping cases and compared it with a greedy approach. 
Given the partial order constraints, invalid routes and invalid 
ordering among routes can be eliminated. Even when there is only one 
overlap in the environment, it will cut off the number of task 
sequences by 50$\%$. In terms of optimality, corollary~\ref{coro:convex} 
continues to hold. Due to limited space, evaluation for the overlapping 
cases is not presented in Section~\ref{sec:evaluation}; we mention that, 
besides less computation cost, the results show no difference from the non-overlapping setting. 
\vspace{-1mm}


\section{Evaluation}\label{sec:evaluation}
\vspace{-1mm}
\def\cvrpgd{\textsc{Alg-GD-CVRP}\xspace}
\def\cvrpgds{\textsc{GD-CVRP}\xspace}
\def\cvrpbcp{\textsc{VRP-Solver}\xspace}
\def\cvrpbcps{\textsc{BCP-CVRP}\xspace}
\def\cvrgodpfh{\textsc{Alg-DP-FH-CVRGO}\xspace}
We carried out extensive simulation studies to evaluate the performance 
of the proposed methods. These efforts are described and discussed in this 
section. The proposed algorithms are implemented in Python and all the 
experiments are executed on an Intel$^\circledR$ Xeon$^\circledR$ CPU at 
3.00GHz.

\vspace{-1mm}
\subsection{Environment Setup}
\vspace{-1mm}
We selected multiple customer location and object weight distributions 
for realistic evaluation. More specifically, problem instances are 
generated in two phases. First, object weights are uniformly randomly 
selected with a lower and upper weight bound. For a given integer $k$, 
three ranges are used: (1) $[0, 1]$, (2) $[1/k, 1]$, and (3) $[1/k, 2/k]$. 
This choice models the practical setting that a lower limit on 
pickup/delivery weight is placed, below which it becomes uneconomical 
to do so. In our experiments, $k = 7$. The $[0, 1]$ case consumes the 
case of $k$ being arbitrarily large. 

After weights are picked, the orientation of the objects are uniformly 
selected in $[0, 2\pi)$. The mass center of the objects in the workspace 
are selected according to two distributions:
\begin{enumerate}
\vspace{-1mm}
\item \textbf{Uniform}: The locations of objects are uniformly 
selected in the bounded 2D workspace. 
\item \textbf{Gaussian}: The locations of objects follow a two-dimensional 
Gaussian distribution where heavy objects are closer to the mean of the 
distribution (Fig.~\ref{fig:setup}(a)). 
\vspace{-1mm}
\end{enumerate}
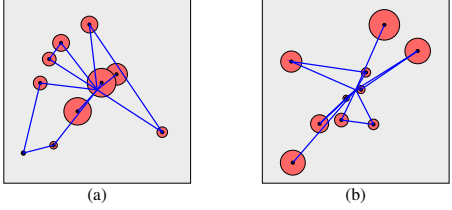
\begin{figure}[ht]
\begin{center}

\resizebox{0.3\columnwidth}{!}{
\begin{tikzpicture}
	  \draw [fill=gray!15] (-3,-3) rectangle (3,3);
    \coordinate (p1) at (-0.25, 2.1);
    \coordinate (p2) at (-1.16, 1.51);
    \coordinate (p3) at (-1.83, 0.22);
    \coordinate (p4) at (0.62, 0.5);
    \coordinate (p5) at (-1.40, -1.78);
    \coordinate (p6) at (-1.54, 0.99);
    \coordinate (p7) at (-2.37, -2.03);
    \coordinate (p8) at (-0.63, -0.69);
    \coordinate (p9) at (2.09, -1.36);
    \coordinate (p10) at (0.14, 0.23);
    \coordinate (d) at (0, 0);

    \filldraw[ fill=red!60](p1) circle (0.27);
    \filldraw[ fill=red!60](p2) circle (0.27);
    \filldraw[ fill=red!60](p3) circle (0.22);
    \filldraw[ fill=red!60](p4) circle (0.35);
    \filldraw[ fill=red!60](p5) circle (0.12);
    \filldraw[ fill=red!60](p6) circle (0.22);
    \filldraw[ fill=red!60](p7) circle (0.07);
    \filldraw[ fill=red!60](p8) circle (0.44);
    \filldraw[ fill=red!60](p9) circle (0.17);
    \filldraw[ fill=red!60](p10) circle (0.46);
    
    \filldraw[ fill=black](p1) circle (0.05);
    \filldraw[ fill=black](p2) circle (0.05);
    \filldraw[ fill=black](p3) circle (0.05);
    \filldraw[ fill=black](p4) circle (0.05);
    \filldraw[ fill=black](p5) circle (0.05);
    \filldraw[ fill=black](p6) circle (0.05);
    \filldraw[ fill=black](p7) circle (0.05);
    \filldraw[ fill=black](p8) circle (0.05);
    \filldraw[ fill=black](p9) circle (0.05);
    \filldraw[ fill=black](p10) circle (0.05);
    
    \draw[line width = 0.4mm,color=blue,-] (d) -- (p9) -- (p1) -- (d);
    \draw[line width = 0.4mm,color=blue,-] (d) -- (p10);
    \draw[line width = 0.4mm,color=blue,-] (d) -- (p4);
    \draw[line width = 0.4mm,color=blue,-] (d) -- (p2) -- (p6) -- (d);
    \draw[line width = 0.4mm,color=blue,-] (d) -- (p8);
    \draw[line width = 0.4mm,color=blue,-] (d) -- (p3) -- (p7) -- (p5) -- (d);
    \node[black] at (0,-3.4){\LARGE{(a)}};
    \end{tikzpicture}
    }
		\hspace{6mm}
\resizebox{0.3\columnwidth}{!}{
\begin{tikzpicture}
	  \draw [fill=gray!15] (-3,-3) rectangle (3,3);
    \coordinate (p1) at (0.17, 0.01);
    \coordinate (p2) at (0.33, 0.56);
    \coordinate (p3) at (1.99, 1.25);
    \coordinate (p4) at (-1.16, -1.09);
    \coordinate (p5) at (-0.31, -0.27);
    \coordinate (p6) at (0.57, -1.11);
    \coordinate (p7) at (0.92, 2.09);
    \coordinate (p8) at (-2.07, 0.91);
    \coordinate (p9) at (-2.02, -2.34);
    \coordinate (p10) at (-0.46, -0.97);
    \coordinate (d) at (0, 0);

    \filldraw[ fill=red!60](p1) circle (0.13);
    \filldraw[ fill=red!60](p2) circle (0.15);
    \filldraw[ fill=red!60](p3) circle (0.41);
    \filldraw[ fill=red!60](p4) circle (0.29);
    \filldraw[ fill=red!60](p5) circle (0.10);
    \filldraw[ fill=red!60](p6) circle (0.17);
    \filldraw[ fill=red!60](p7) circle (0.49);
    \filldraw[ fill=red!60](p8) circle (0.34);
    \filldraw[ fill=red!60](p9) circle (0.40);
    \filldraw[ fill=red!60](p10) circle (0.22);
    
    \filldraw[ fill=black](p1) circle (0.05);
    \filldraw[ fill=black](p2) circle (0.05);
    \filldraw[ fill=black](p3) circle (0.05);
    \filldraw[ fill=black](p4) circle (0.05);
    \filldraw[ fill=black](p5) circle (0.05);
    \filldraw[ fill=black](p6) circle (0.05);
    \filldraw[ fill=black](p7) circle (0.05);
    \filldraw[ fill=black](p8) circle (0.05);
    \filldraw[ fill=black](p9) circle (0.05);
    \filldraw[ fill=black](p10) circle (0.05);
    
    \draw[line width = 0.4mm,color=blue,-] (d) -- (p1) -- (p3) -- (d);
    \draw[line width = 0.4mm,color=blue,-] (d) -- (p9);
    \draw[line width = 0.4mm,color=blue,-] (d) -- (p7);
    \draw[line width = 0.4mm,color=blue,-] (d) -- (p10) -- (p6) -- (d);
    \draw[line width = 0.4mm,color=blue,-] (d) -- (p5) -- (p4) -- (d);
    \draw[line width = 0.4mm,color=blue,-] (d) -- (p2) -- (p8) -- (d);
    \node[black] at (0,-3.4){\LARGE{(b)}};
    \end{tikzpicture}
    }
\vspace{-2mm}
\end{center}
\caption{\label{fig:setup}Gaussian (a) and inverse-Gaussian (b) distributions 
of customer locations. The size of the disc signifies the objects' size and/or weights.}
\vspace{-2mm}
\end{figure}

We selected the Gaussian setup to model the collapsing of a large
object into multiple smaller pieces (e.g., after an explosion), 
where it is likely for heavy pieces to be close to the epicenter. 
During the evaluation, we further examined the ``inverse'' 
Gaussian setting where heavy objects are more likely to be away 
from the Gaussian mean (Fig.~\ref{fig:setup}(b)). We omit the result
for this setting as it demonstrates performance 
characteristics similar to the uniform setting. 

\vspace{-2mm}
\subsection{Performance Evaluation of \cvrg}
\vspace{-2mm}
We evaluate the performance of four algorithms on \cvrg: 
(1) \cvrgdps, an exact algorithm based on \cvrpdp, with tour costs computed with \eia, 
(2) \cvrgdpfhs, the finite horizon version of \cvrgdps, with $h=10$,
(3) \cvrggds, a greedy best-first algorithm that picks the closest 
object satisfying capacity constraints, 
(4) \cvrgbcps, which uses a state-of-the-art branch-cut-and-price method 
\cvrpbcps \cite{pessoa2019generic} to compute a solution with centroid 
of the objects and then use \eia to shorten the paths. We also include 
\cvrpbcps, to see \eia's contribution. 

\begin{figure}[h!]
\vspace{1mm}
    \centering
    \begin{overpic}[width={\iftwocolumn 0.3\columnwidth \else 2.2in \fi},tics=5]
{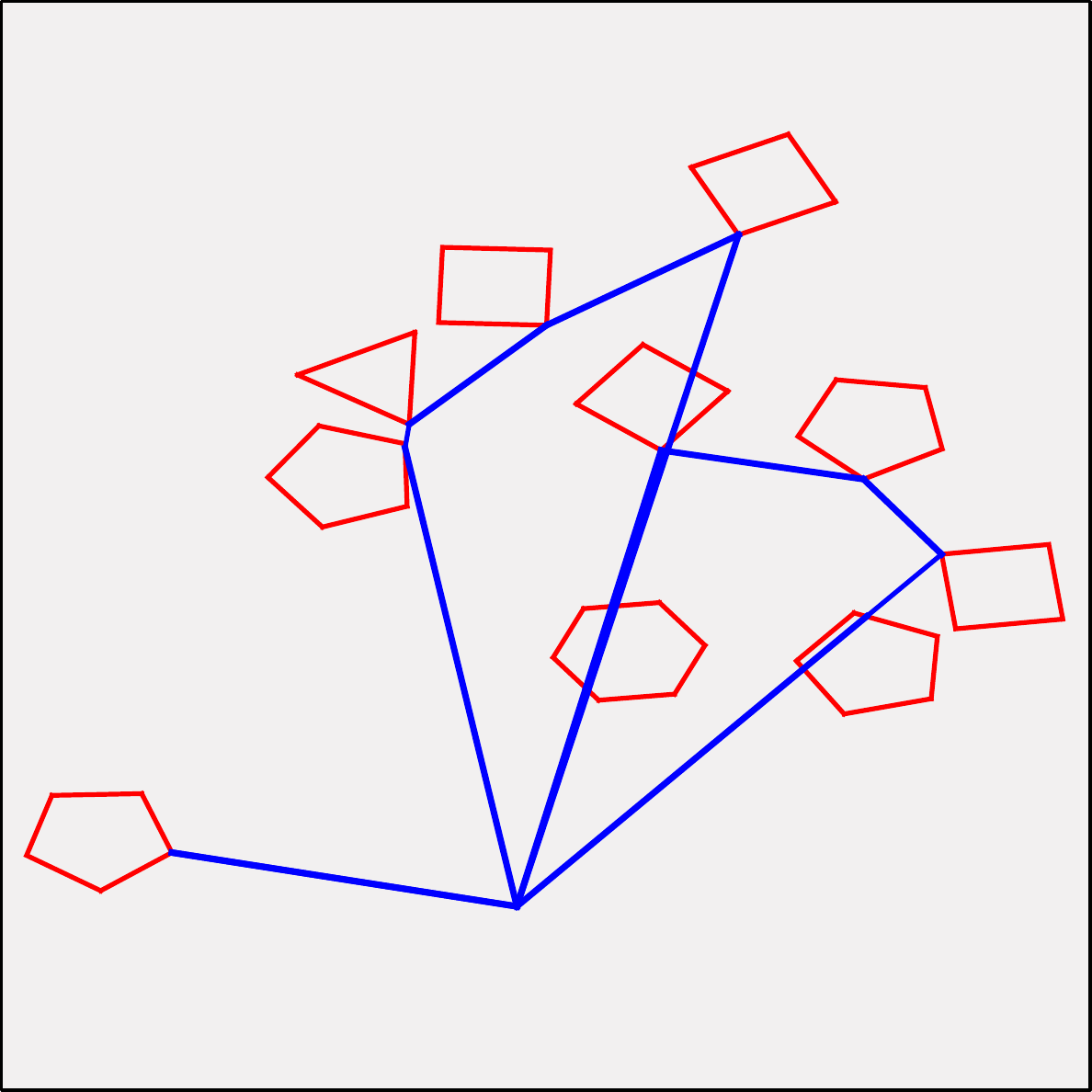}
\end{overpic}
\hspace{6mm}
    \begin{overpic}[width={\iftwocolumn 0.3\columnwidth \else 2.2in \fi},tics=5]
{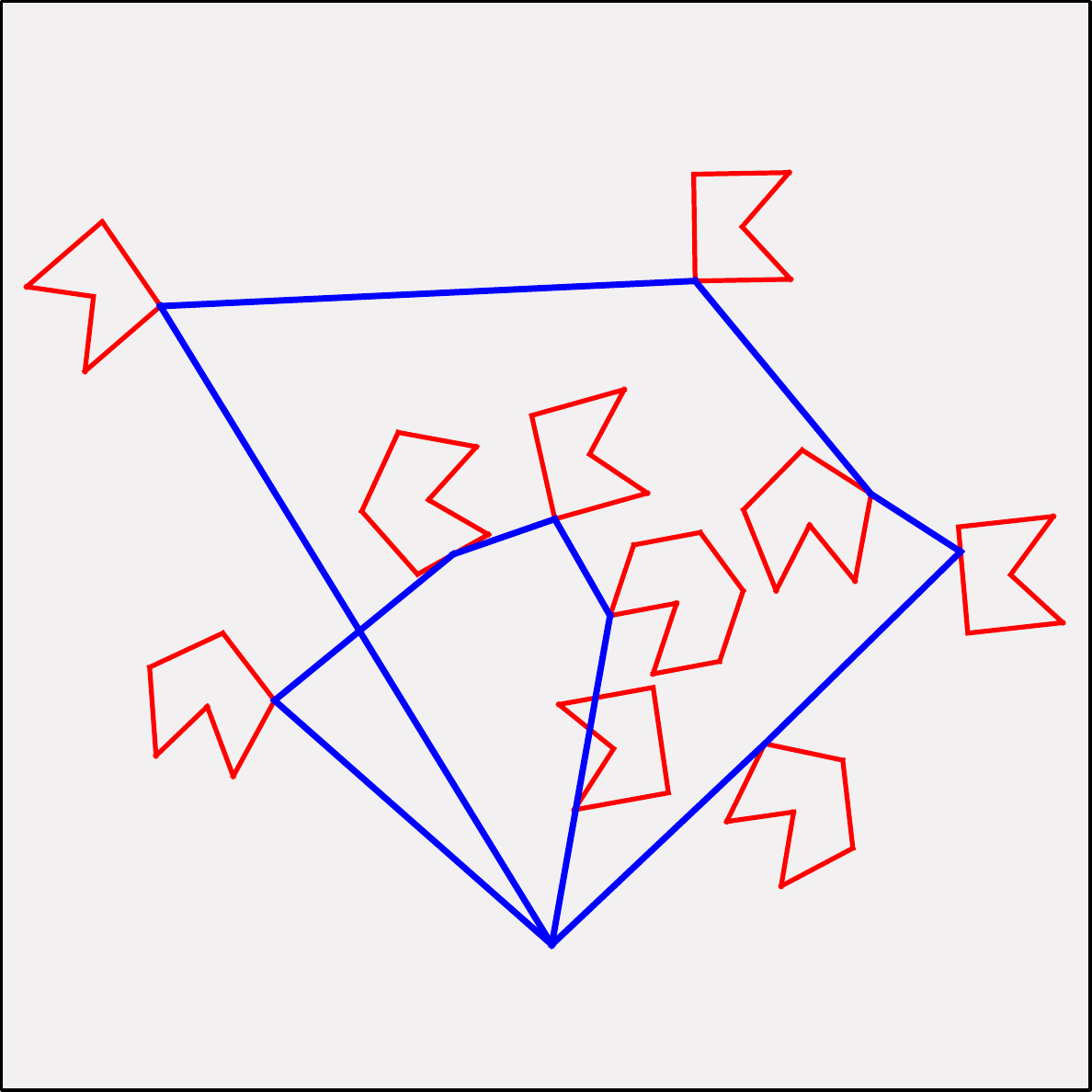}
\end{overpic}
\vspace*{1mm}
    \caption{\label{fig:cvrgExample}(left) A (convex polygon only) \cvrg 
		instance with an optimal solution generated by \cvrgdps. (right) A 
		\cvrg instance with non-convex polygons and an optimal
		solution.}
\vspace{-1mm}
\end{figure}

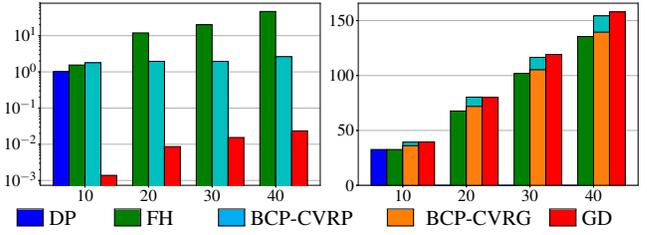
\begin{figure}[h!]
\begin{center}
\begin{overpic}[width={\iftwocolumn \columnwidth \else 2.1in \fi},tics=5]{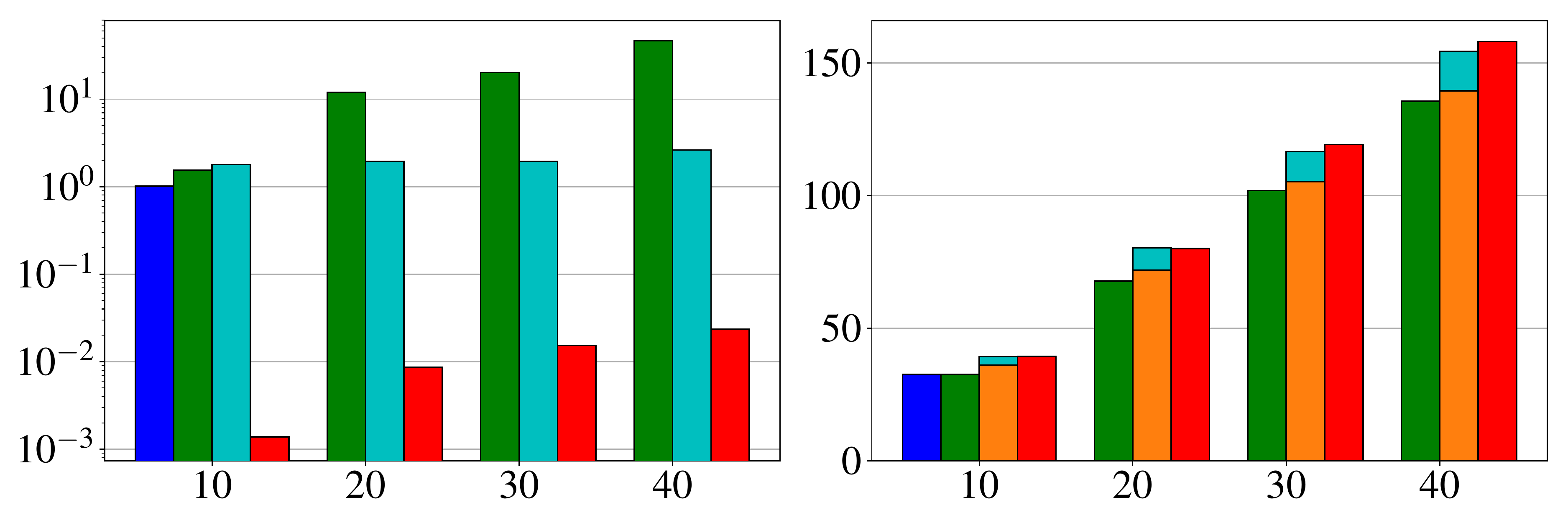}
\put(3,-1){\fcolorbox{black}{blue}{  }}
\put(8, -2.15
){{\footnotesize DP}}
\definecolor{Green}{RGB}{0, 128, 0}
\put(18,-1){\fcolorbox{black}{Green}{  }}
\put(23, -2.15
){{\footnotesize FH}}
\put(34,-1){\fcolorbox{black}{cyan}{  }}
\put(39, -2.15
){{\footnotesize BCP-CVRP}}
\put(60,-1){\fcolorbox{black}{orange}{  }}
\put(66, -2.15
){{\footnotesize BCP-CVRG}}
\definecolor{Red}{RGB}{255, 0, 0}
\put(85,-1){\fcolorbox{black}{Red}{  }}
\put(90, -2.15
){{\footnotesize GD}}
\end{overpic}
\end{center}
\vspace*{1mm}
\caption{\label{fig:cvrg-1} Algorithm performance for \cvrg under uniform 
customer location distribution with 10-40 customers (left: time (s); right: optimality (unitless)). 
BCP-CVRP and BCP-CVRG are overlaid with the 
shorter bar in front. For computational time, BCP variants take 
essentially the same amount of time as \eia is fast. }
\end{figure}

Our evaluation models aerial delivery applications (Fig.~\ref{fig:drone}), 
where each customer location is a polygon. A typical solution is 
illustrated in Fig.~\ref{fig:cvrgExample} (the left figure), where the 
blue lines are the tours. 
In generating the polygonal regions for evaluation, we made 
the diameter of the polygon bounded by $\frac{1}{10}$ of the workspace side 
length. The result is given in Fig.~\ref{fig:cvrg-1}. The solution produced 
by \cvrgbcps is not optimal and is about $5\%$ worse than the solution produced 
by \cvrgdpfhs, which computes near optimal solution (\cvrgdps computes optimal 
solution when \eia is optimal). 

\begin{figure}[h!]
\begin{center}
\begin{overpic}[width={\iftwocolumn \columnwidth \else 3.45in \fi},tics=5]{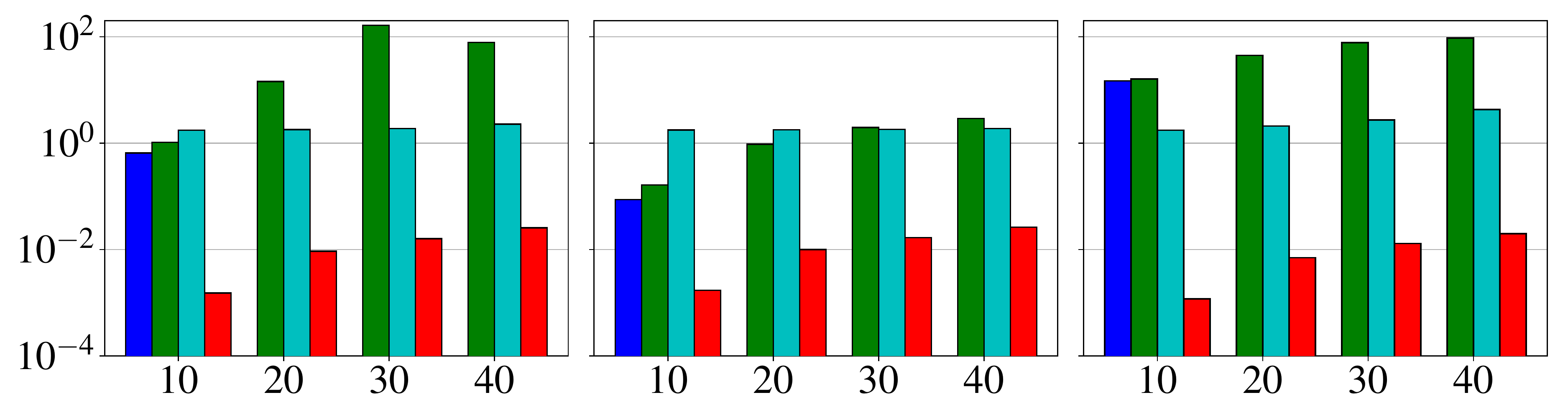}
\put(3,-1){\fcolorbox{black}{blue}{  }}
\put(8, -2.15
){{\footnotesize DP}}
\definecolor{Green}{RGB}{0, 128, 0}
\put(18,-1){\fcolorbox{black}{Green}{  }}
\put(23, -2.15
){{\footnotesize FH}}
\put(34,-1){\fcolorbox{black}{cyan}{  }}
\put(39, -2.15
){{\footnotesize BCP-CVRP}}
\put(60,-1){\fcolorbox{black}{orange}{  }}
\put(66, -2.15
){{\footnotesize BCP-CVRG}}
\definecolor{Red}{RGB}{255, 0, 0}
\put(88,-1){\fcolorbox{black}{Red}{  }}
\put(93, -2.15
){{\footnotesize GD}}
\end{overpic}
\end{center}
\vspace*{1mm}
\caption{\label{fig:cvrg-2-time} Algorithm running time in seconds for \cvrg 
under Gaussian distribution with 10-40 customers for different weight distributions: [left] $[0, 1]$.
[middle]$[1/k,1]$. [right] $[1/k, 2/k]$.}
\end{figure}

\begin{figure}[h!]
\vspace{-0.5mm}
\begin{center}
\begin{overpic}[width={\iftwocolumn \columnwidth \else 3.45in \fi},tics=5]{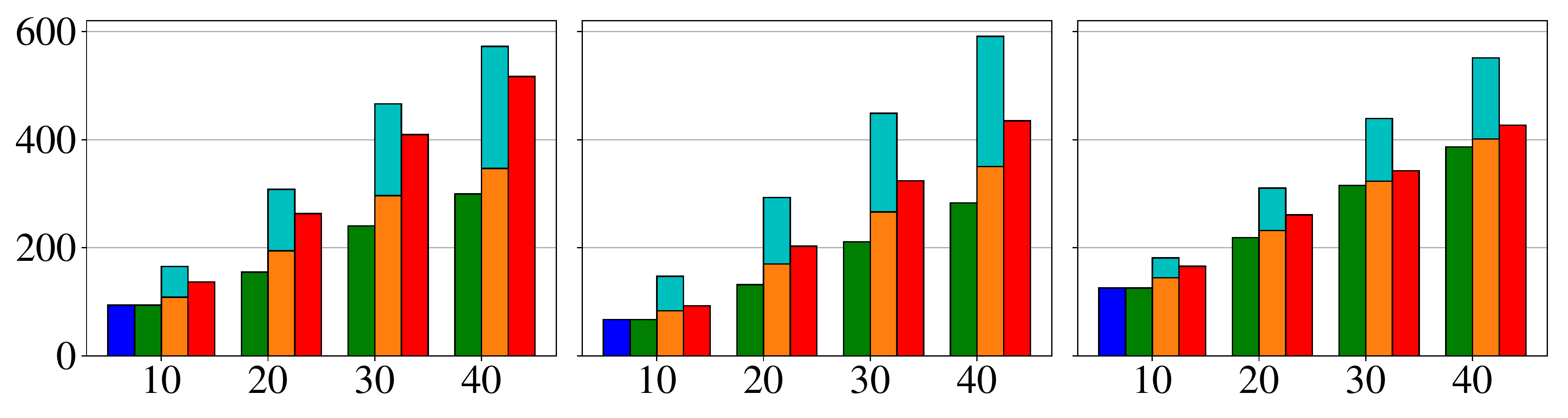}
\put(3,-1){\fcolorbox{black}{blue}{  }}
\put(8, -2.15
){{\footnotesize DP}}
\definecolor{Green}{RGB}{0, 128, 0}
\put(18,-1){\fcolorbox{black}{Green}{  }}
\put(23, -2.15
){{\footnotesize FH}}
\put(34,-1){\fcolorbox{black}{cyan}{  }}
\put(39, -2.15
){{\footnotesize BCP-CVRP}}
\put(60,-1){\fcolorbox{black}{orange}{  }}
\put(65, -2.15
){{\footnotesize BCP-CVRG}}
\definecolor{Red}{RGB}{255, 0, 0}
\put(88,-1){\fcolorbox{black}{Red}{  }}
\put(93, -2.15
){{\footnotesize GD}}
\end{overpic}
\end{center}
\vspace*{1mm}
\caption{\label{fig:cvrg-2-cost} Corresponding solution quality for 
cases in Fig.~\ref{fig:cvrg-2-time}.}
\vspace{-1mm}
\end{figure}
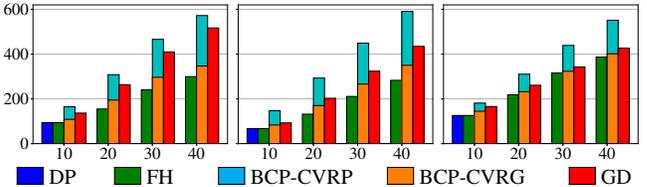

The second evaluation examines the Gaussian setup with all three weight 
distributions. The running time and solution quality results are given 
in Fig.~\ref{fig:cvrg-2-time} and Fig.~\ref{fig:cvrg-2-cost}, respectively.
We also observe that \cvrgdpfhs runs much faster as we lower bound the object 
weight, as expected in practice. In terms of solution optimality, 
\cvrgdpfhs is near optimal and significantly outperforms the greedy method 
(by $50\%$) as well as \cvrgbcps (by $15\%$ to 
$20\%$ percent). This shows superiority of \cvrgdpfhs in solution quality.

We evaluated performance of \cvrgdps and \cvrgdpfhs for cases where
customer regions are non-convex. For the case where there are $10$ 
regions, we compute the optimal solution by splitting each non-convex
region into multiple convex ones and then run a modified version of 
\cvrgdps. The result confirms that \cvrgdps and \cvrgdpfhs both compute 
the same as the exact optimal solution. A typical case is illustrated 
in Fig.~\ref{fig:cvrgExample} (the right figure). The evaluation 
empirically suggests that \cvrgdpfhs is expected to compute high-quality 
solution even when regions are non-convex.

In addition to comprehensive numerical studies, we carried out physics 
based simulations of the aerial delivery scenario in the Unreal Engine to 
obtain a more realistic estimate of the travel and delivery time. The setup 
is similar to that illustrated in Fig.~\ref{fig:drone} and Fig.~\ref{fig:cvrgExample}, 
with $10$ to 
$20$ objects. Over $10$ runs of different setups for $10$ objects, the time 
cost ratio between \cvrgdps/\cvrgdpfhs, \cvrgbcps, and \cvrggds is 1:1.07:1.17 
(for this case, recall that \cvrgdps and \cvrgdpfhs compute the same 
optimal solutions). For $20$ objects, the ratio between \cvrgdpfhs, \cvrgbcps, 
and \cvrggds is 1:1.02:1.16. Considering the time that is required for 
making deliveries, these results largely agree with the earlier numerical results. 

\begin{figure}[h!]
\centering
    \begin{overpic}[width={\iftwocolumn 0.9\columnwidth \else 2.5in \fi},tics=5]{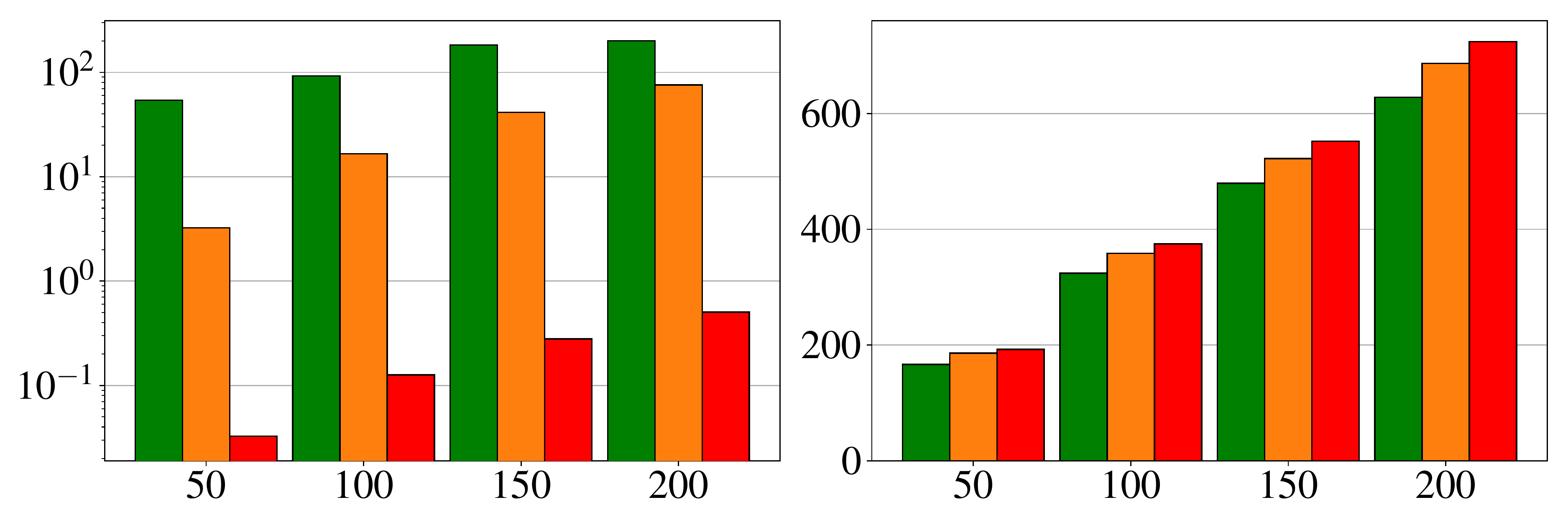}
    \definecolor{Green}{RGB}{0, 128, 0}
\put(14,-1){\fcolorbox{black}{Green}{  }}
\put(20, -2.15
){{\footnotesize FH}}
\put(39,-1){\fcolorbox{black}{orange}{  }}
\put(46, -2.15
){{\footnotesize \cvrgbcps}}
\put(72,-1){\fcolorbox{black}{Red}{  }}
\put(78, -2.15
){{\footnotesize GD}}
\end{overpic}
\vspace*{1.5mm}
    \caption{ Algorithm performance for CVRG drone delivery with large number of customers(50-200).}
    \label{fig:l-delivery}
\end{figure}
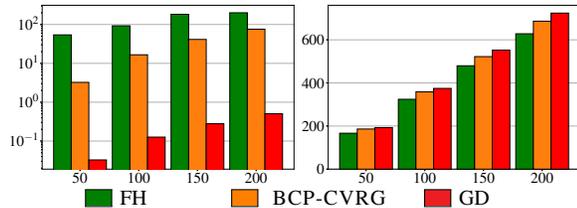

Our methods for \cvrg have excellent scalability. In under $200$ seconds, 
\cvrgbcps and \cvrgdpfhs readily scale to over 200 objects. Outcome of the 
computational experiments are given in Fig.~\ref{fig:l-delivery}, in which 
uniform object distribution with uniform weight distribution in $[0, 1]$ is 
used. 


We observe that the trend agrees with earlier results. In particular, 
for \cvrg, \cvrgdpfhs does considerably better 
than both \cvrgbcps ($\sim 10\%$) and the greedy method ($\sim 20\%$).

\begin{figure}[h!]
\centering
    \begin{overpic}[width={\iftwocolumn 0.9\columnwidth \else 2.5in \fi},tics=5]{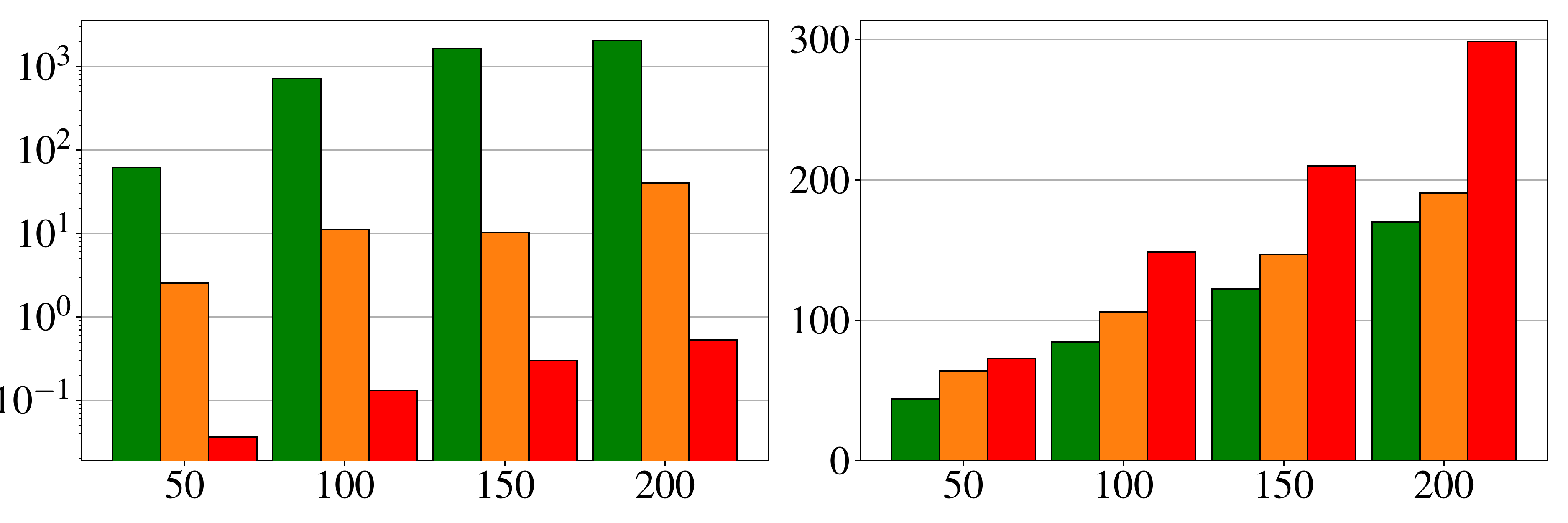}
    \definecolor{Green}{RGB}{0, 128, 0}
\put(14,-1){\fcolorbox{black}{Green}{  }}
\put(20, -2.15
){{\footnotesize FH}}
\put(39,-1){\fcolorbox{black}{orange}{  }}
\put(46, -2.15
){{\footnotesize \cvrgbcps}}
\put(72,-1){\fcolorbox{black}{Red}{  }}
\put(78, -2.15
){{\footnotesize GD}}
\end{overpic}
\vspace*{+2mm}
    \caption{ Algorithm performance for CVRG drone delivery under the Gaussian 
		distribution with large number of customers(50-200), both object locations and weights are uniformly distributed.}
    \label{fig:l-delivery-Gaussian}
\end{figure}
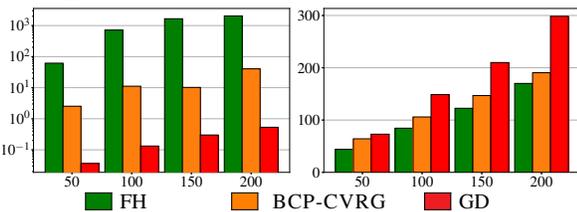

As a last evaluation, we attempted large scale instances for \cvrg where the regions are 
chosen following the Gaussian distribution. As shown in 
Fig.~\ref{fig:l-delivery-Gaussian}, the trend of the solution quality 
agrees with previous experiments under Gaussian distribution. On the
other hand, the running time for \cvrgdpfhs is comparatively large. 
This can be attributed to the regions with low weights tending to be the 
last delivery targets. When there are hundreds of regions in the instance, 
most of the candidates in the last rounds of DP process are with low 
weights. This causes the amount of candidates in the later tours to be disproportionately large, 
requiring more time to go through. 

\vspace{-4mm}

\section{Conclusion}\label{sec:conclusion}
\vspace{-1mm}
In this work, we examined \cvrg as a \cvrp variant where the target 
locations have non-trivial geometry, with applications toward a 
variety of robotics tasks including aerial delivery, rescue, clutter 
removal, and so on.
Solving \cvrg optimally requires the careful selection of exact customer 
locations for completing pickup or delivery tasks. We developed multiple 
efficient algorithms for tackling \cvrg, and evaluated their performance 
as the number of customers, customer geometry, and customer distribution 
changes. In all cases, it was shown that our algorithms provide significant 
advantage in computing high quality solutions as compared with greedy 
approaches and \cvrp solver based methods.

{\footnotesize
\bibliographystyle{IEEEtran}
\bibliography{bib}
}

\end{document}